\documentclass[12pt]{article}
\usepackage{amsmath,epsfig,amssymb,amsfonts,amsthm,epsfig,verbatim,color,lscape,longtable,natbib,epstopdf,subfigure,graphicx,ltxtable,dcolumn,array,multirow,booktabs,adjustbox,url,mathrsfs,algorithm,algorithmic,appendix,adjustbox,rotating,tablefootnote,latexsym,amsmath}

\usepackage{setspace}
\usepackage{fullpage}
\usepackage{hyperref}
\hypersetup{colorlinks = true,
	linkcolor  = black,
	anchorcolor = black,
	citecolor = black}
\usepackage{arydshln}

\newtheorem{Def}{Definition}
\newtheorem{Con}{Condition}

\newtheorem{Mth}{Theorem}

\newtheorem{Rem}{Remark}

\numberwithin{equation}{section}
\newtheorem{prop}{Proposition}

\def\argmin{\mathop{\rm argmin}}

\def\diag{\mathop{\rm diag}}
\def\pr{\mathop{\rm pr}}

\newcommand{\indep}{\perp \!\!\! \perp}

\newcommand {\bfSigma} {\mbox{\boldmath $\Sigma$}}
\newcommand {\bfDelta} {\mbox{\boldmath $\Delta$}}

\newcommand {\bfTheta} {\mbox{\boldmath $\Theta$}}

\newcommand {\bfepsilon} {\mbox{\boldmath $\epsilon$}}
\newcommand {\bfOmega} {\mbox{\boldmath $\Omega$}}

\def\bx{\boldsymbol{x}}
\def\bX{\boldsymbol{X}}

\def\bB{\boldsymbol{B}}
\def\bA{\boldsymbol{A}}

\def\be{\boldsymbol{e}}
\def\bI{\boldsymbol{I}}

\title{Local structural transfer learning of non-Gaussian DAG and its application to multi-source cancer gene regulatory networks}
\author{
	Mingyang Ren$^{1}$, Xin He$^{2}$ and Junhui Wang$^{3}$ \\
	$^1$School of Mathematical Sciences, Shanghai Jiao Tong University \\ 
    $^2$School of Statistics and Data Science, \\
    Shanghai University of Finance and Economics \\
     $^3$ Department of Statistics, The Chinese University of Hong Kong
    }
\date{}

\begin{document}
	
	\maketitle
	
    \onehalfspacing
    \begin{abstract}
    \noindent Directed acyclic graph (DAG) is pivotal for modeling gene regulatory networks in cancer pathogenesis, yet single-study genomic data is often limited for accurate DAG reconstruction. Although multi-source datasets provide augmented sample sizes, substantial heterogeneity persists across real-world cohorts. It remains an open question how to pool the heterogeneous data with only local similarity together for better DAG structure reconstruction in the target study. In this paper, we introduce a novel set of structural similarity measures for DAGs and then present a transfer DAG learning framework by effectively leveraging information from auxiliary DAGs with node-level local similarities. Our theoretical analysis shows substantial improvement in terms of DAG reconstruction in the target study, which is in sharp contrast to most existing transfer learning methods requiring global similarity. The advantage of the proposed transfer DAG learning is also supported by extensive numerical experiments on both synthetic data and multi-source cancer genomics data.

    \end{abstract}
	
    \doublespacing
   \section{Introduction}
   
   Directed gene regulatory networks (GRNs) explicitly map how numerous genes activate or repress one another. Understanding these precise directional interactions is fundamental for deciphering cancer disease mechanisms, and identifying key drivers within these networks is essential for developing effective targeted therapies aimed at specific molecular vulnerabilities.  To construct, analyze, and interpret complex directional dependencies inherent in GRNs, Directed Acyclic Graphs (DAGs), specifically designed to represent a series of nodes (genes) connected by directed edges (regulatory influences), are an indispensable statistical framework.
   It has also attracted tremendous interest from researchers, and various DAG learning algorithms have been proposed, including \cite{chen2019causal} and \cite{yuan2019constrained}, on Gaussian DAG with equal noise variance \citep{peters2014identifiability}. The pioneering work of \cite{shimizu2006linear} unveiled the research route on non-Gaussian DAG, which can be free from the identifiability conditions required in Gaussian DAGs, capitalizing on the property of non-Gaussian distribution, and has received increasing attention in recent years \citep{hyvarinen2013pairwise, wang2020high, zhao2022learning}, 
   
   As the motivating examples, our analysis goal is to construct directed GRNs of the specific {\it rare T-cell subgroup of non-small cell lung cancer} and the {\it rare subtype of breast cancer}, whose sample size can be extremely limited due to the rarity of diseases and the high acquisition costs. However, existing DAG learning methods, despite their successes, have mainly focused on a dataset from one single cancer study and neglected the possibility of relevant data collected to contribute to a given target institution. 
   To fully utilize the heterogeneous data from different sources, transfer learning, aiming at transferring the information from auxiliary domains to improve learning performance on the target domain, is a promising solution with gaining attention. Transfer learning has been extensively investigated in the machine learning community; interested readers may refer to \cite{zhuang2020comprehensive} and the references therein. Yet, most existing transfer learning methods are algorithm-based, and the statistical guarantees of transfer learning still remain insufficient and grow in popularity. Particularly, \cite{cai2021transfer} proposed some minimax and adaptive transfer learning-based classifiers. \cite{li2022transferb} proposed a transfer learning method for high-dimensional linear regression models with minimax optimality, which is extended to high-dimensional generalized linear regression models \citep{tian2022transfer,li2023estimation} and federated learning \citep{li2021targeting}. In contrast, transfer learning for unsupervised learning is in its infancy  \citep{li2022transfera}, and largely follows the idea of parameter transfer as for supervised learning.

   Two types of multi-source cancer genomics data reflecting the macro-level heterogeneity of populations and the micro-level heterogeneity of cells considered in this article present two outstanding domain-specific challenges for transfer DAG learning that cannot be properly addressed by directly extending the existing transfer learning methods. First, in directed GRNs, the primary scientific objective is often to identify which directional regulatory relationships exist (edges) rather than just quantifying their strength (parameters). Recovering graph structure through {\it structural transfer} fundamentally differs from {\it parameter transfer}, requiring new appropriate similarity measures, globally or locally, that capture topological relationships rather than just numerical proximity. 
   Second, while {\it global or overall similarity} is a widely adopted assumption in the existing literature of transfer learning, which assures that all parameters of an informative auxiliary domain are close to those of the target domain as a whole, real multi-source cancer data consistently exhibit only local structural similarities. Our exploratory analysis of non-small cell lung cancer T-cell subgroups reveals this stark contrast: while the CD4-C5-EOMES target subgroup shares specific edges like E2F3→STK4 with the CD4-C3-GNLY auxiliary subgroup, their overall network topologies differ substantially (Figure \ref{eg-NSCLC}a). Similarly, in breast cancer subtypes, triple-negative breast cancer (TNBC) shares regulatory relationships like DVL1→HRAS with Luminal B but exhibits divergent global structures (Figure \ref{eg-NSCLC}b). This local similarity paradigm is further complicated by three factors: asymmetric sharing, where an auxiliary domain may contain relevant sub-structures absent in the target domain; context-specificity, where identical genes may participate in different regulatory roles across cancer types; and partial overlap, where similarity exists only within specific functional modules but not others. These nuances render global similarity assumptions untenable for most cancer genomics applications, demanding new methodologies capable of detecting and transferring local structural knowledge without requiring identical global structures.

   In this article, in addition to parametric similarity, we propose a novel set of structural similarity measures for auxiliary and target DAGs, from global-level, layer-level to node-level. Then, focusing on linear non-Gaussian DAGs, we further propose a transfer DAG learning framework, which can detect structure-informative auxiliary DAGs benefiting from the well-defined structural similarity measures, and the topological layers of the target DAG can be better recovered with the help of the detected auxiliary DAGs. The theoretical and numerical improvements of the proposed transfer DAG learning framework compared with its single-task learning alternatives are also investigated and demonstrated in analyses of the rare T-cell subgroup of non-small cell lung cancer and the rare subtype of breast cancer.

   The main contributions of the proposed framework are multi-fold. First, a novel set of DAG structural similarity measures with different levels are proposed, which are the progressive and comprehensive definition development from global and macroscopic to local and microcosmic. Such structural similarity measures constitute a fundamental complement to the parametric similarity in transfer learning.
   Second, a transfer DAG learning framework is developed accordingly, which accommodates auxiliary DAGs with different levels of structural similarities, achieving both {\it structure transfer} and {\it parameter transfer}. In the realistic case where the structure-informative auxiliary DAGs are only locally similar to the target DAG on some nodes, structure transfer with only local similarity is an important extension of the existing parameter transfer paradigm, requiring global or overall similarity. 
   Third, the theoretical analysis shows that the proposed transfer DAG learning is advantageous in terms of DAG reconstruction of the target domain even in high-dimensional settings, including more relaxed conditions, reduced probability of failure in topology layer reconstruction and target sample complexity, and improved estimation.
   Last but not least, the proposed transfer DAG learning is applied on
   two types of multi-source cancer genomics data, including multiple T-cell subgroups of non-small cell lung cancer and multiple subtypes of breast cancer, which provide some interesting biological insights from the directed regulatory relationships between genes.
   

   \section{Multi-source cancer genomics data with local similarity}\label{real-ex}
   
   Multi-source heterogeneous data are highly prevalent in the biomedical field. This heterogeneity manifests at different levels, including the macro-level heterogeneity of populations, such as different subtypes of the same disease, and the micro-level heterogeneity of cells, such as distinct subgroup structures of immune cells within an individual with the same disease. The two real-world data sets analyzed in this paper involve heterogeneity issues at both micro and macro levels.
   
   At the micro level, we consider single-cell RNA sequencing (scRNA-seq) data from non-small cell lung cancer (NSCLC) patients. NSCLC represents a major global health challenge, accounting for approximately 85\% of all lung cancer cases and serving as a leading cause of cancer-related mortality. The advent of scRNA-seq has unprecedentedly enabled the resolution of complex T-cell heterogeneity within the tumor microenvironment, facilitating the exploration of the potential immune functions of distinct T-cell subgroups, thereby revolutionizing NSCLC research. Understanding this diversity is biologically crucial for deciphering immune evasion mechanisms, predicting immunotherapy responses, and identifying novel therapeutic targets. Previous studies have reported T-cell heterogeneity at the gene expression level, identifying several subgroups among 9,055 T cells from NSCLC patients \citep{guo2018Global}. Further exploration of the directed gene regulatory networks within specific subgroups will deepen the understanding of T-cell heterogeneity.

   \begin{figure}[!htb]
   	\centering
   	\subfigure{\includegraphics[scale = 0.425]{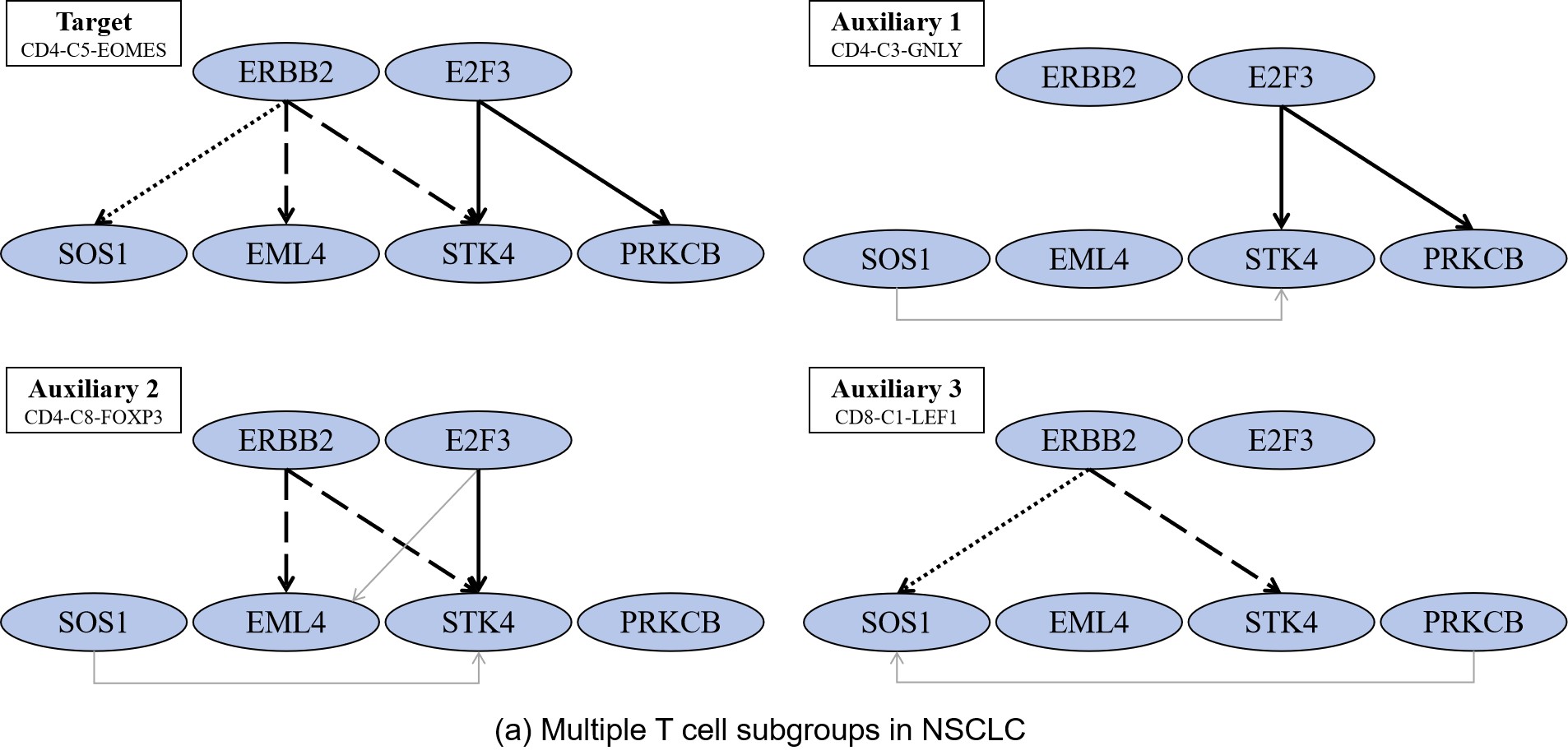}} \\
   	\subfigure{\includegraphics[scale = 0.425]{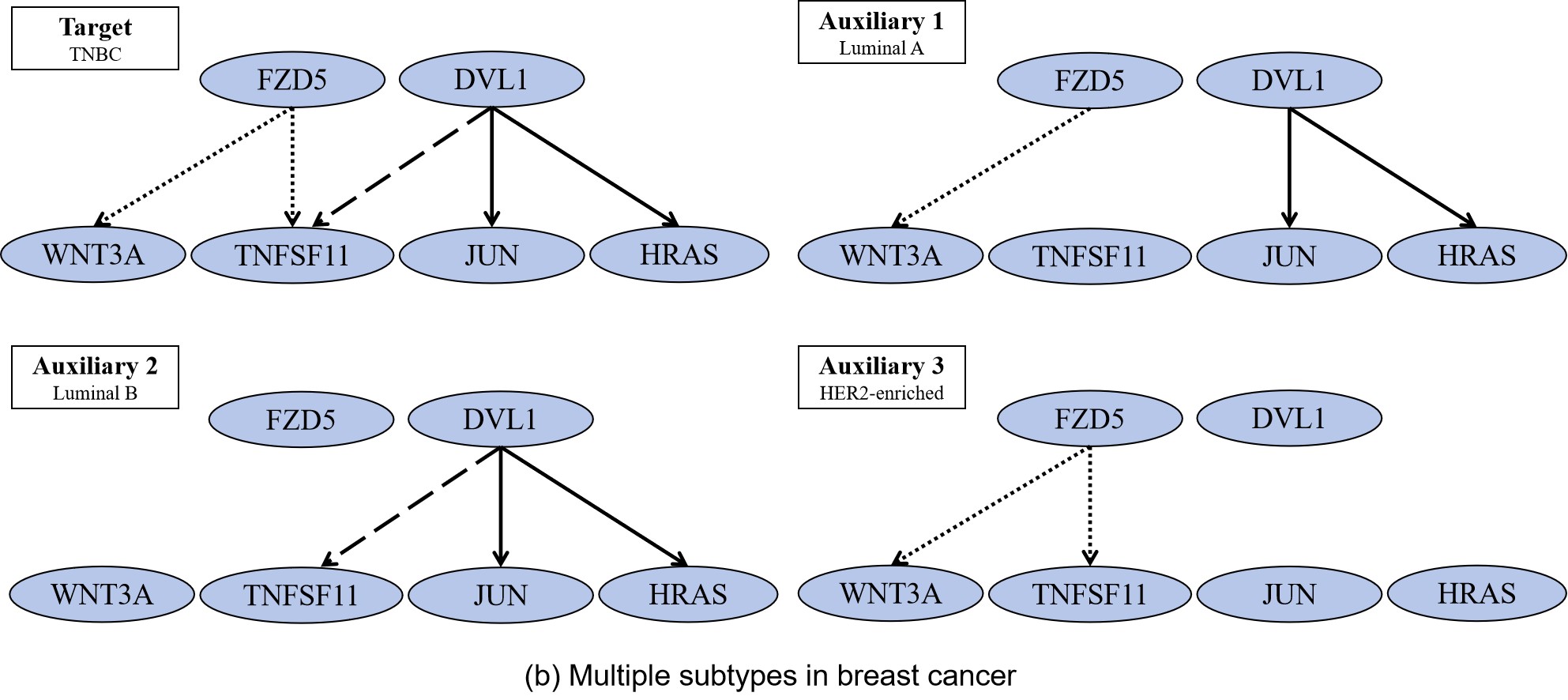}} 
   	\caption{The local DAG structures of multiple T-cell subgroups of NSCLC (upper) and multiple subtypes of breast cancer (below). Each distinct edge type in the target domain corresponds to a different auxiliary domain sharing that specific edge structure.}
   	\label{eg-NSCLC}
   \end{figure}
   
   Heterogeneity between subgroups is also reflected in their gene regulatory networks. This renders existing methods, which rely on the assumption of identical DAG structures in different domains, inapplicable to such problems. For instance, considering the CD4-C5-EOMES subgroup (161 cells), a key activated CD4+ T cell state, as the target domain and three auxiliary domains, CD4-C3-GNLY (433 cells), CD4-C8-FOXP3 (427 cells), and CD8-C1-LEF1 (303 cells), we exploratively used single-DAG learning method of \cite{zhao2022learning} to estimate their respective DAG structures, aiming to characterize the similarity between the four DAGs roughly.
   Figure \ref{eg-NSCLC}a displays representative local structures involving 6 genes within these four DAGs. Clearly, no auxiliary DAG possesses a structure identical to the target DAG. Fortunately, each auxiliary DAG exhibits local structural similarities to the target DAG. For example, Auxiliary DAG1 shares the edges E2F3→STK4 and E2F3→PRKCB with the target domain; Auxiliary DAG2 additionally shares ERBB2→EML4 and ERBB2→STK4; and Auxiliary DAG3 further shares ERBB2→SOS1. This validates the existence of local structural similarities within the real-world data. Moreover, the non-Gaussianity is also very common in scRNA-seq data, due to significant biases in the expression levels of different genes in different cells.
   
   At the macro level, we consider gene expression data from population cohorts representing four breast cancer subtypes, Luminal A, Luminal B, HER2-enriched, and TNBC. Using TNBC, a relatively rare subtype with a notably poor prognosis, as the target domain, the other three types as the auxiliary domains, we illustrate the local structures of the four DAGs in Figure \ref{eg-NSCLC}b. The local similarity of DAGs can also be observed.
   
   These observations profoundly demonstrate that, from the perspective of DAG learning, real-world multi-source heterogeneous cancer omics data exhibit characteristics that far exceed the assumption of overall similarity. Instead, local structural similarity proves to be a more realistic and prevalent phenomenon. This insight strongly motivates us to propose a novel measure of local structural similarity for heterogeneous DAGs and to develop the corresponding novel method to address this practical challenge.

   \section{DAG similarity measures}
   
   To provide a clear foundation for the proposed multiple levels of DAG similarity measures in Section \ref{sec_sim}, some necessary notations and basic concepts of DAG that will be used throughout the paper are first introduced.

   \subsection{Preliminaries}
   
   \textbf{Notations.} Denote $\| \boldsymbol{u} \|_q$ as the  $l_q$-norm of a vector $\boldsymbol{u}$, for $q \geqslant 0$. For a matrix $\bA = (A_{ij})_{1 \leqslant i,j \leqslant p}$, let $\bA_{j}$ be its $j$-th column and $\bA_{-jj}$ be the $\bA_{j}$ without $A_{jj}$, $\| \bA \|_{q, \infty} = \max_{1 \leqslant j \leqslant p} \| \bA_{j} \|_q$, $\| \bA \|_1 = \sum_{j=1}^{p} \| \bA_{j} \|_1$, $\| \bA \|_{\max} = \max_{1 \leqslant i,j \leqslant p} | A_{ij} |$, $\| \bA \|_{1, \mathrm{off}} = \sum_{1 \leqslant i \neq j \leqslant p} |A_{ij}|$, and $\| \bA \|_F$ be the Frobenius norm of $\bA$. When $\bA$ is symmetric, we further denote $\psi_{\min}(\bA)$ and $\psi_{\max}(\bA)$ as the smallest and largest eigenvalues of $\bA$, respectively. Let $\bA_{S}$ be the submatrix of $\bA$ with both rows and columns labeled by $S$. Let $|S|$ be the cardinality of a set $S$ and $[K] = \{ 1, \cdots, K \}$ be the $K$-set for any positive integer $K$. For sequences $a_n$ and $b_n$, define $a_n \lesssim b_n$ if there exists a positive constant $C$ such that $a_n \leqslant C b_n$, and $ a_n\asymp b_n$ if $a_n \lesssim b_n$ and $b_n \lesssim a_n$. For two real numbers $a$ and $b$, define $a \wedge b = \min \{ a, b\}$ and $a \vee b = \max \{ a, b\}$. The superscript $^*$ on a parameter marks its true value.

   \noindent \textbf{DAG model}. Consider a DAG ${\cal G}=\{ {\cal N},{\cal E}\}$, encoding the joint distribution $P(\bx)$ of $\bx=(x_1,...,x_p)^T \in \mathbb{R}^p$, where  ${\cal N}=\{1,\ldots,p\}$ consists of a set of nodes associated with each coordinate of $\bx$, and  $ {\cal E}\subset {\cal N}\times{\cal N}$ consists of all the directed edges among the nodes. For each node $j$, its parents, children, and descendants are denoted as $\mbox{pa}_j$, $\text{ch}_j$, and $\text{de}_j$, respectively. Each node $x_j$ is also centered with mean zero. 
   Suppose that the joint distribution of $\bx$ can also be embedded into a linear structural equation model (SEM),
   \begin{align}\label{eqn:1}
   	x_j=\sum_{l \in \mbox{pa}_j} {\beta_{jl}} x_l + \epsilon_j; \ j=1,...,p,
   \end{align}
   where ${\beta_{jl}}\neq 0$ for any $l \in \mbox{pa}_j$,  $\epsilon_j$ denotes a continuous {\it non-Gaussian} noise with variance $\sigma_j^2$, and $\epsilon_l \indep \epsilon_j$ for any $l \neq j$. This independent noise condition also implies that $\epsilon_j \indep x_l$ for any $l \notin \mbox{de}_j\cup \{j\}$. 
   Furthermore, the SEM model in \eqref{eqn:1} can be rewritten as a matrix form $\bx=\bB\bx+{\bfepsilon}$ with ${\bB}=(\beta_{jl})_{j,l}\in \mathbb{R}^{p\times p}$ and $\bfepsilon=(\epsilon_1,...,\epsilon_p)^{\top}$ is the noise vector with covariance matrix $\mathbf{\Omega}=\diag \{ \sigma_1^2,...,\sigma_p^2 \}$. It follows that $\bx =(\bI-{\bB})^{-1}\bfepsilon ={\bA}\bfepsilon$,
   and $x_j=\sum_{l=1}^p a_{jl}\epsilon_l$, where $a_{jl}$ is  the $(j,l)$-th element of $\bA=(\bI-{\bB})^{-1}$, representing the total effect of  $x_l$ on $x_j$. Moreover, the covariance matrix of $\bx$ is $\bfSigma = (\bI-\bB)^{-1} \bfOmega (\bI-\bB)^{-T}$, and the corresponding precision matrix is 
   $\bfTheta=\bf\Sigma^{-1}=(\bI-\bB)^{T} \bfOmega^{-1}(\bI-\bB)$.
   We remark that non-Gaussian DAGs do not require equal or ordered noise variance conditions to ensure identifiability, which is necessary for Gaussian DAGs \citep{peters2014identifiability, ghoshal2018learning} and may parameterize structural information.
   
   \noindent \textbf{Topological layers}. Any DAG can be converted into a unique topological layer structure \citep{zhao2022learning}. Without loss of generality, assume ${\cal G}$ has a total of $T$ topological layers. All the leaf nodes and isolated nodes in ${\cal G}$ belong to the lowest layer ${\cal A}_{0}$. For $t = 1, \ldots, T-1$, ${\cal A}_t$ denotes all the nodes contained in the $t$-th layer, whose longest distance to a leaf node is exactly $t$.  It is then clear that  $\cup_{t=0}^{T-1}{\cal A}_{t}={\cal N}$. Further, for each node $j \in {\cal A}_t$, it follows from the layer construction that $\mbox{pa}_j \subset {\cal S}_{t+1}=\cup_{d=t+1}^{T-1}{\cal A}_d$, and ${\cal S}_0={\cal N}$. Therefore, acyclicity is automatically guaranteed. Figure \ref{para-struc} illustrates a toy DAG in panel (a), and its converted topological structure with four layers in panel (b).
   Denote $\mathbf{\Theta}^{{\cal S}_t}$ and $\mathbf{\Sigma}_{{\cal S}_t}$ as the precision and covariance matrices of $\bx_{{\cal S}_t}$, respectively. Note that 
   $\mathbf{\Theta}^{{\cal S}_{0}}= \mathbf{\Theta}$. The topological layers can help reconstruct the DAG structure. Particularly, for any $j \in {\cal S}_t$, regressing $x_j$ on $\bx_{{\cal S}_t \setminus \{j\}}$ yields the expected residual $e_{j, {\cal S}_t}=x_j + \bx_{{\cal S}_t \setminus \{j\}}^{\top} [\mathbf{\Theta}^{{\cal S}_t}]_{-jj}/[\mathbf{\Theta}^{{\cal S}_t}]_{jj}$. It was shown in \cite{zhao2022learning} that 
   \begin{align}\label{detest}
   	j \in {\cal A}_{t} \text{ if and only if } e_{j, {\cal S}_t}\indep x_l \text{ for any } l\in { {\cal S}_t\backslash \{j\} }.
   \end{align}
   This result leads to a simple bottom-up learning algorithm to re-construct ${\cal A}_{t}$ for $t = 0,1, \ldots, T-1$ by using the independence test. More importantly, this layer reconstruction algorithm also motivates us to design new concepts and methods for transfer learning of DAG accordingly.

   \subsection{Multiple levels of DAG similarity}\label{sec_sim}
   
   Denote the target DAG as ${\cal G}_0=\{ {\cal N},{\cal E}_0\}$, which encodes the joint distribution $P_0(\bx)$ of $\bx=(x_1,...,x_p)^T$ with the precision matrix $\bfTheta$. Suppose that there also exist $K$ auxiliary DAGs ${\cal G}_k=\{ {\cal N},{\cal E}_k\}$ with $k \in [K]$, encoding the joint distribution $P_k(\bx^{(k)})$ of $\bx^{(k)}=(x_1^{(k)},...,x_p^{(k)})^T$ with the precision matrix $\bfTheta^{(k)}$ and the covariance matrix $\bfSigma^{(k)}$. The auxiliary domain $k$ can also be embedded into a linear SEM \eqref{eqn:1} with different parameters including the continuous non-Gaussian noise $\epsilon^{(k)}_j$ and its variance $[\sigma_j^{(k)}]^2$. It is of great interest to improve the learning accuracy of ${\cal G}_0$ in the target domain by leveraging information from ${\cal G}_k$ in the auxiliary domains.
   
   The effectiveness of such a transfer learning task is heavily based on measuring the similarity between ${\cal G}_0$ and ${\cal G}_k$. In particular, we propose two similarity measures for DAG, corresponding to the presence or  strength of causal effects between nodes in different domains as measured in the parameters or graph structures. 
   
   \subsubsection{Structural Similarity}
   
   Definition \ref{sim-struc} is among the first attempts to provide some well-defined and comprehensive structural similarities between DAGs, which take advantage of the topological layers and shall greatly facilitate transfer learning for DAGs in various scenarios.
   
   \begin{Def}[Structural Similarity]\label{sim-struc}
   	
   	\begin{itemize}
   		\item[(a)] ${\cal G}_k$ is {\rm global structure-informative}, if it has the same number of layers as ${\cal G}_0$, and ${\cal A}_t = {\cal A}_t^{(k)}$ for any $t = 0, 1, \cdots, T-1$, where ${\cal A}_t$ and ${\cal A}_t^{(k)}$ are the nodes contained in the $t$-th layers of ${\cal G}_0$ and ${\cal G}_k$, respectively.
   		\item[(b)] ${\cal G}_k$ is {\rm layer-$t$ structure-informative}, if for some $t \in \{ 0, 1, \cdots, T-1 \}$, there exists $t^{\prime} \geqslant 0$, such that ${\cal A}_t = {\cal A}_{t^{\prime}}^{(k)}$ and ${\cal S}_t = {\cal S}_{t^{\prime}}^{(k)}$.
   		\item[(c)] ${\cal G}_k$ is {\rm node-$j$ structure-informative}, if (i) $\text{de}_j^{(k)} \cap {\cal S}_{t_j} = \emptyset$ and (ii) $\text{de}_j^{(k)} \cap {\cal A}_{t_j - 1} \ne \emptyset$ with $t_j \geqslant 1$, where $t_j$ denotes the layer to which node-$j$ belongs in ${\cal G}_0$, and $\text{de}_j^{(k)}$ denotes node $j$'s descendants in ${\cal G}_k$.
   	\end{itemize}
   \end{Def}
   
\begin{figure}[!htb]
   	\centering
   	\includegraphics[scale = 0.425]{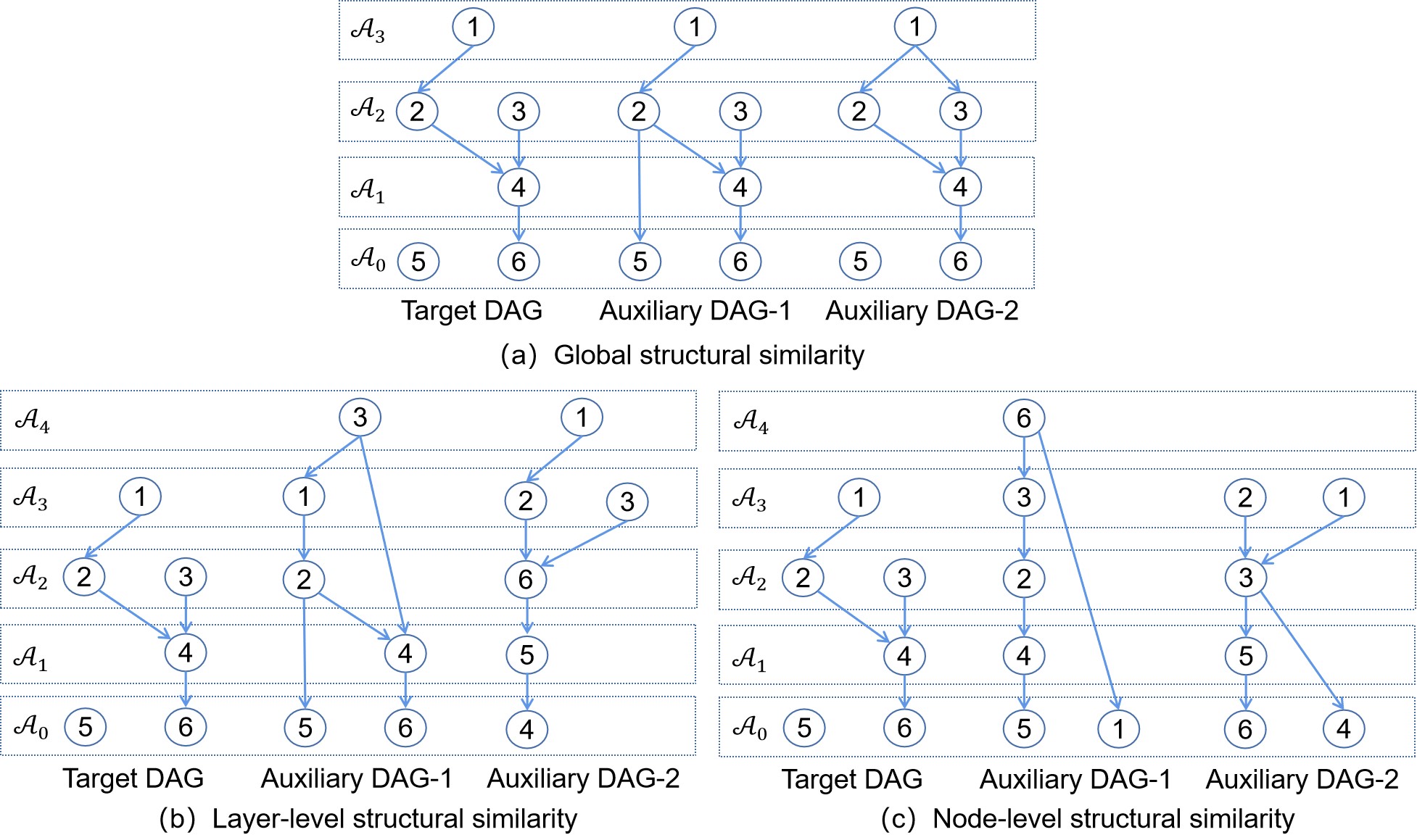}
   	\caption{Some toy examples for DAGs in the target and auxiliary domains with different levels of structural similarities.}
   	\label{struc-fig}
   \end{figure}

   From Definition \ref{sim-struc}a, a global structure-informative DAG needs to have exactly the same topological layers as the target DAG. It is important to remark that the global structural similarity concerns only the membership of nodes in each topological layer, and does not impose any requirements on edges between nodes in different layers. The layer-level structural similarity in Definition \ref{sim-struc}b focuses on the relative positions of one particular topological layer in ${\cal G}_0$ and ${\cal G}_k$. This similarity ensures that information from ${\cal G}_k$ can be efficiently transferred when constructing the topological layer of interest in ${\cal G}_0$. The node-level structural similarity in Definition \ref{sim-struc}c is a very weak requirement, focusing on local and microcosmic structures of DAGs. A node-$j$ structure-informative DAG only shares a similar relative position of node $j$ as ${\cal G}_0$, where the relative position can be depicted in two cases. Case (i) requires no overlap between node $j$'s descendants in ${\cal G}_k$ and ${\cal S}_{t_j}$ in ${\cal G}_0$, and thus information from ${\cal G}_k$ can prevent assigning node $j$ to a topological layer higher than $t_j$ in ${\cal G}_0$. Case (ii) requires node $j$'s descendants in ${\cal G}_k$ overlap with ${\cal A}_{t_j - 1}$ in ${\cal G}_0$, which can prevent assigning node $j$ to a topological layer lower than $t_j$ in ${\cal G}_0$. These three levels of structural similarities are progressive. 
   For illustration, Figure \ref{struc-fig} gives some toy examples of these three structural similarities, where only two auxiliary DAGs are informative while others can be of arbitrary structure and thus non-informative.
   
   It can be observed in Figure \ref{struc-fig}a that both ${\cal G}_1$ and ${\cal G}_2$ are global structure-informative, which have the same topological layers as ${\cal G}_0$ but their directed edge sets are all different. In Figure \ref{struc-fig}b, ${\cal G}_1$ is layer-$0$ and layer-$1$ structure-informative, and ${\cal G}_2$ is layer-2 and layer-3 structure-informative. They share the same nodes in the respective layers as ${\cal G}_0$, but the composition of other layers is largely different. In Figure \ref{struc-fig}c, ${\cal G}_1$ is node-2, node-4, and node-5 structure-informative, and ${\cal G}_2$ are node-1, node-3, and node-6 structure-informative. Take node-2 as an example, when constructing layer ${\cal A}_1$ of ${\cal G}_0$, ${\cal G}_1$ provides useful information to prevent mis-assigning node 2 to ${\cal A}_1$, since node 4 is a child of node 2 in ${\cal G}_1$ and belongs to ${\cal A}_1$ of ${\cal G}_0$. On the other hand, when constructing ${\cal A}_2$ of ${\cal G}_0$ after ${\cal A}_0$ and ${\cal A}_1$ are determined, ${\cal G}_1$ provides information to assign node 2 to ${\cal A}_2$, thanks to the fact that  node 2's descendants in ${\cal G}_1$ do not appear in ${\cal S}_2$ of ${\cal G}_0$. Similar observations can also be made for other nodes.

   \subsubsection{Parametric Similarity}
   Denote the divergence matrix as $\bfDelta^{(k)}_{{\cal S}_t} = \bfTheta^{{\cal S}_t} \bfSigma^{(k)}_{{\cal S}_t} - \boldsymbol{I}_{|{\cal S}_t|}$, where $\boldsymbol{I}_{|{\cal S}_t|}$ is the $|{\cal S}_t|$-dimensional identity matrix. It measures the similarity between the precision matrices regarding the $t$-th layer in the target and the $k$-th auxiliary domains. By using $\bfDelta^{(k)}_{{\cal S}_t}$, Definition \ref{sim-para} defines a parametric similarity between ${\cal G}_0$ and ${\cal G}_k$.
   
   \begin{Def}[Parametric Similarity]\label{sim-para}
   	${\cal G}_k$ is {\rm layer-$t$ parameter-informative}, if for some small $h_t>0$, $\|  \bfDelta^{(k)}_{{\cal S}_t} \|_{1, \infty} + \|  ( \bfDelta^{(k)}_{{\cal S}_t} )^{\top} \|_{1, \infty} \leqslant h_t$.
   \end{Def}
   
\begin{figure}[!htb]
   	\centering
   	\includegraphics[scale = 0.5]{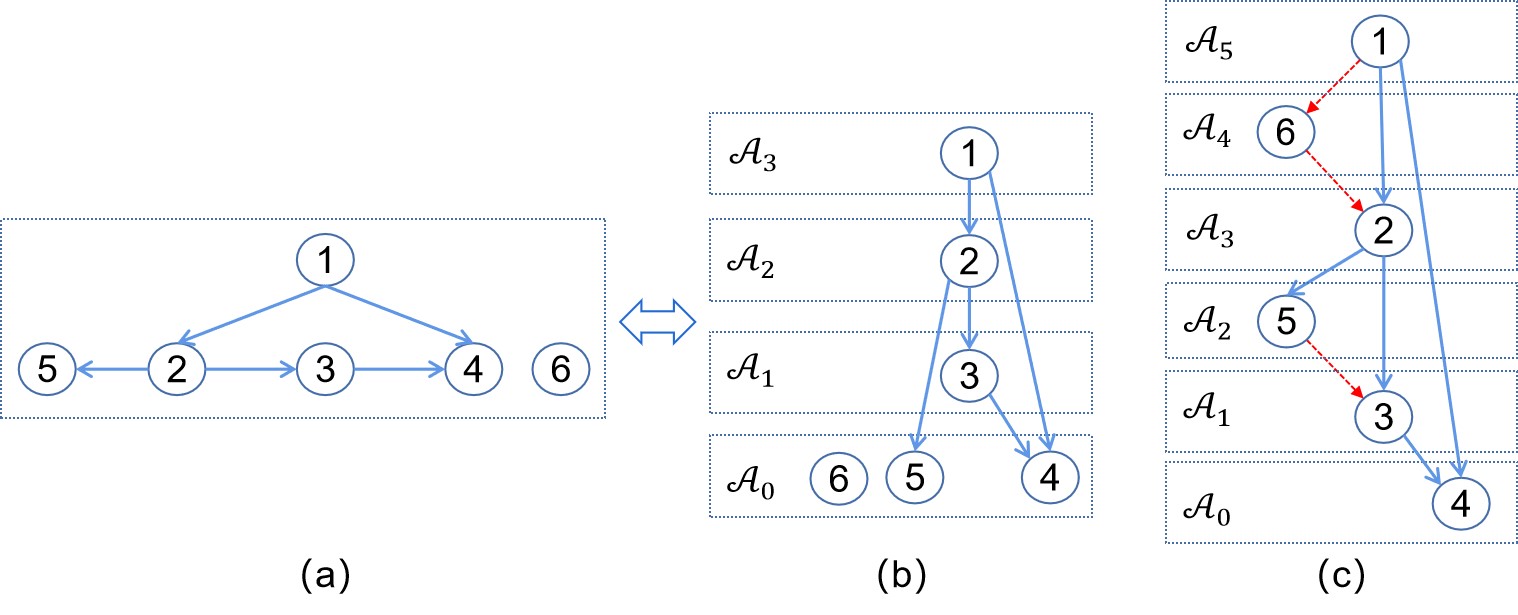}
   	\caption{A toy DAG in (a) and its converted topological structure with four layers in (b), and (c) gives another DAG sharing parametric similarity but not structural similarity with that in (b).}
   	\label{para-struc}
   \end{figure}

   It is clear that a small value of $h_t$ indicates the precision matrices corresponding to ${\cal S}_t$ in the target and auxiliary domains are close, and thus the information from the auxiliary domain can help improve the estimation accuracy of $\bfTheta$ in the target domain. Definition \ref{sim-para} is reminiscent of the parameter similarity of Gaussian graphical models in literature \citep{li2022transfera}, but the parametric similarity in $\bfTheta$ does not necessarily imply similar DAG structures. Figure \ref{para-struc} gives a toy example, where the DAGs in (b) and (c) are parametric similar, provided that the three red-directed edges in (c) are of weak strength. Yet, the graph structures and topological layers in (b) and (c) are substantially different, and thus additional structural similarity is necessary to quantify the similarity between DAGs. 
   

   \begin{Rem}
   	The structure recovery is the core task of DAG learning, and this paper mainly focuses on transfer learning under structural similarity. The parametric similarity in Definition \ref{sim-para} can help improve the estimation of edge strengths, but it is not necessary for the recovery of the target DAG structures, which is often of more importance in DAG learning. More theoretical discussions on the scenarios without parametric similarity are provided in Section \ref{property}.
   \end{Rem}

   \begin{Rem}\label{glo_com}
   	The global structural similarity shares the same spirit with the overall similarity in multi-task DAG learning, which jointly estimates multiple DAGs from different datasets \citep{shimizu2012joint, Wang2020joint}, such as the consistent causal order across tasks \citep{chen2021multi}. Another related problem is to estimate the same DAG shared by multiple different datasets, where the nodes of these datasets may overlap but are not necessarily the same \citep{danks2008integrating,triantafillou2015constraint,mooij2020joint}. It is clear that all these problems rely on the overall similarity, and their implications and utilization are substantially different from the transfer DAG learning considered here.
   \end{Rem}

   \section{Transfer DAG learning}\label{methods}
   
   Suppose in addition to $\bX =\ ( {\bx}_{1}, \ldots, {\bx}_{p} ) \in \mathbb{R}^{n\times p}$ with ${\bx}_j=(x_{1j},...,x_{nj})^{\top}$ from the target domain with ${\cal G}_0$, we also observe $\{\bX^{(k)} \}_{k=1}^K$ from $K$ auxiliary domains with $\bX^{(k)} \in \mathbb{R}^{n_k \times p}$, where $n$ and $n_k$ are sample sizes of the target and the $k$-th auxiliary domains, respectively, and $n \lesssim n_k$ for any $k \in [K]$. The primary goal is to leverage information in $\{\bX^{(k)} \}_{k=1}^K$ to help construct ${\cal G}_0$.

   \subsection{Proposed generic framework}
   {The proposed generic transfer DAG learning framework is summarized in Algorithm~\ref{alg:2}. In this framework, the topological layers $\mathcal{A}_t$ are constructed in a bottom-up fashion from $t=0$ to $t=T-1$, and directed edges between nodes in different layers are then determined based on the estimated target precision matrix. Specifically, Algorithm~\ref{alg:2} consists of two main steps for constructing each layer $\mathcal{A}_t$. First, the target DAG is used to detect locally structure-informative auxiliary DAGs for each node by leveraging the well-defined structural similarities introduced in Section~\ref{sec_sim}. Second, a majority voting scheme is adopted to aggregate the information from the detected auxiliary DAGs. The technical details are provided in Section~\ref{case2}.}
   It should be noted that in all steps, parameter-informative auxiliary DAGs are not necessary, while if one wishes to utilize them to enhance the estimation of edge strengths, well-established parameter transfer methods can be introduced, as shown in Section \ref{mpt}.

   \begin{algorithm}[htb]
   	\caption{Transfer Learning Framework for DAG}
   	\label{alg:2}
   	\begin{algorithmic}[1]
   		\STATE \textbf{Input: } $\bX$, $\{ \bX^{(k)} \}_{k=1}^{K}$, $\widehat{\cal S}_t=\{1,...,p\}$, $\widehat{\bB}=\{\hat{\beta}_{ij}\}_{i,j=1}^p=\mathbf{0}_{p\times p}$, and $t=0$.
   		\STATE  {\bf Repeat:} until   $|\widehat{\cal S}_t| \leqslant 1$:
   		\begin{itemize}
   			\setlength\itemsep{-0.5em}
   			\item[(a)]  Estimate $\widehat{\mathbf{\Theta}}^{\widehat{\cal S}_t}$ via well-established graphical model methods.
   			\item[(b)]  Construct $\widehat{\cal A}_{t}$ via independence test leveraging the detected informative auxiliary domains;
   			\item[(c)]  Estimate $\widehat{\mbox{pa}}_{j}=\{l \in \widehat{\cal S}_t\backslash \widehat{\cal A}_t: [\widehat{{\bfTheta}}^{\widehat{\cal S}_t}]_{jl}\neq 0 \}$ and $\widehat{\beta}_{jl}=-[\widehat{{\bfTheta}}^{\widehat{\cal S}_t}]_{jl}/[\widehat{{\bfTheta}}^{\widehat{\cal S}_t}]_{jj}$ for any $j \in \widehat{\cal A}_{t}$ and $l \in \widehat{\mbox{pa}}_j$;
   			\item[(d)] Update $\widehat{\cal S}_{t+1} \leftarrow \widehat{\cal S}_t \setminus \widehat{\cal A}_{t}$ and $t\leftarrow t+1$.
   		\end{itemize}
   		\STATE	{ If} $|\widehat{\cal S}_t|=1$,  set $\widehat{T}=t+1$ and $\widehat{\cal A}_{t}=\widehat{\cal S}_t$; otherwise set $\widehat{T}=t$.
   		\STATE	{\bf Return:} $\{\widehat{\cal A}_t\}_{t=0}^{\widehat{T}-1}$ and  $\widehat{\bB}$.
   	\end{algorithmic}
   \end{algorithm}
   
   The definition of node-level structural similarity essentially characterizes the identity of the microstructure at the node level between the auxiliary and target DAGs, implying consistent independence relationships between other nodes and the expected residuals corresponding to a node. Through the connection of Equation \eqref{detest}, such local structural similarity can further be reflected in the independence measure. Therefore, it is utilized to detect structure-informative auxiliary DAGs and then to conduct independence tests to establish ${\cal A}_{t}$'s. In Step 2b of Algorithm \ref{alg:2}, the distance covariance and its related independence test \citep{szekely2007measuring} are employed. To be self-contained, we briefly restate its definition here. Given $\widehat{\cal S}_t \subseteq \{1,...,p\}$, the distance covariance between the expected residual $e_{j,\widehat{\cal S}_t}$ and the variable $x_l$ is defined as
   $\mbox{dcov}^2(e_{j,\widehat{\cal S}_t}, x_l)=I_{jtl,1} + I_{jtl,2} -2I_{jtl,3}$,
   where $
   I_{jtl,1}={{E}}[|e_{j,\widehat{\cal S}_t} - {e}'_{j,\widehat{\cal S}_t}||x_l - {x}'_l|],\
   I_{jtl,2}={E}[|e_{j,\widehat{\cal S}_t} - {e}'_{j,\widehat{\cal S}_t}|] {E}[|x_l - {x}'_l|]\ \mbox{and} \
   I_{jtl,3}={E} \big ( {E}[|e_{j,\widehat{\cal S}_t} - {e}'_{j,\widehat{\cal S}_t}||e_{j,\widehat{\cal S}_t}] {E}[|x_l - {x}'_l||x_l]\big )
   $ with ${e}'_{j,\widehat{\cal S}_t}$ and ${x}'_l$ denoting an independent copy of ${e}_{j,\widehat{\cal S}_t}$ and $x_l$, respectively. It provides a neat independence criterion in that $\mbox{dcov}^2(e_{j,\widehat{\cal S}_t}, x_l)= 0$ if $e_{j,\widehat{\cal S}_t} \indep x_l$ and $\mbox{dcov}^2(e_{j,\widehat{\cal S}_t}, x_l) > 0$ otherwise.
   On the target domain, an empirical version of $\mbox{dcov}^2(e_{j,\widehat{\cal S}_t}, x_l)$ is 
   \begin{equation}\label{dcov-def}
   	\widehat{\mbox{dcov}}^2 (\widehat{\be}_{j,\widehat{\cal S}_t}, {\bx}_l) = \widetilde{I}_{jtl,1} + \widetilde{I}_{jtl,2} -2\widetilde{I}_{jtl,3},
   \end{equation}
   where $\widehat{\be}_{j,\widehat{\cal S}_t}=\bX_{\widehat{\cal S}_t} [\widehat{\bfTheta}^{\widehat{\cal S}_t}]_{\cdot j}/ [\widehat{\bfTheta}^{\widehat{\cal S}_t}]_{jj} =(\widehat{e}_{1j,\widehat{\cal S}_t} ,..., \widehat{e}_{nj,\widehat{\cal S}_t} )^{\top}$, $\widetilde{I}_{jtl,1}=\frac{1}{n^2} \sum_{i,i^{\prime}=1}^{n} |\widehat{e}_{ij,\widehat{\cal S}_t} - \widehat{e}_{i^{\prime}j,\widehat{\cal S}_t} | |x_{il}-x_{i^{\prime}l}|$, $\widetilde{I}_{jtl,2}=(\frac{1}{n^2} \sum_{i,i^{\prime}=1}^{n} |\widehat{e}_{ij,\widehat{\cal S}_t}-\widehat{e}_{i^{\prime}j,\widehat{\cal S}_t}|) (\frac{1}{n^2} \sum_{i,i^{\prime}} |x_{il}-x_{i^{\prime}l}|)$, and
   $\widetilde{I}_{jtl,3}=\frac{1}{n^3} \sum_{i,i^{\prime},h=1}^{n} |\widehat{e}_{ij,\widehat{\cal S}_t}-\widehat{e}_{hj,\widehat{\cal S}_t}||x_{i^{\prime}l}-x_{hl}|$. It can be employed to test ${\cal H}_{0}: {e}_{j,\widehat{\cal S}_t} \indep x_l \  ~\mbox{vs}~ \ {\cal H}_{1}: e_{j,\widehat{\cal S}_t} \not\!\perp\!\!\!\perp x_l$ with test statistic 
   $\widehat{T}(\widehat{{\be}}_{j,\widehat{\cal S}_t}, {\bx}_l) = \widehat{\mbox{dcov}}^2 (\widehat{{\be}}_{j,\widehat{\cal S}_t}, {\bx}_l) / \widetilde{I}_{jtl,2}$.
   Specifically, ${\cal H}_{0}$ is rejected if $n \widehat{T} \big (\widehat{{\be}}_{j,\widehat{\cal S}_t}, {\bx}_l \big ) > (\Phi^{-1}(1-\alpha/2))^2$, where $\Phi(\cdot)$ denotes the distribution function of standard normal distribution, and $\alpha$ is the significance level. Furthermore, on the $k$-th auxiliary domain, $\widehat{\mbox{dcov}}^2 (\widehat{\be}_{j,\widehat{\cal S}_t}^{(k)}, {\bx}_l^{(k)})$ and $\widehat{T}(\widehat{{\be}}_{j,\widehat{\cal S}_t}^{(k)}, {\bx}_l^{(k)})$ can be similarly defined by replacing $\bX_{\widehat{\cal S}_t}$ and $\widehat{\bfTheta}^{\widehat{\cal S}_t}$ with $\bX_{\widehat{\cal S}_t}^{(k)}$ and $(\widehat{\bfTheta}^{(k)})^{\widehat{\cal S}_t}$, respectively. Note that the
   distance covariance is a nonparametric measure that relaxes distributional assumptions by operating on pairwise distances rather than raw values and is able to detect arbitrary dependence, linear or nonlinear, by testing whether the joint characteristic function factorizes, which is a necessary and sufficient condition for independence.
   The distance covariance relies only on the geometry of ranks and distances, making it robust to outliers and applicable to any distribution with finite first moments. Its test statistic is computed directly from data and typically evaluated via permutation, eliminating the need for parametric assumptions about the underlying data-generating process. Interested readers are referred to \cite{szekely2007measuring} for more details.

   It is clear that a key to transfer DAG learning is to obtain $\widehat{\cal A}_t$ based on the detected auxiliary domains with structural similarities.
   We focus on a more realistic case where no auxiliary DAG is global structure-informative, but node-level structure-informative DAGs satisfying Definition \ref{sim-struc}c for each node are present. The case with global or layer-level structure-informative ${\cal G}_k$'s can be treated in a similar way and thus omitted. 
   
   \subsection{Local structural transfer}\label{case2}
   {Constructing $\mathcal{A}_t$ via local structural transfer consists of two core steps. The first step is detection of node-level structure-informative auxiliary DAGs for each node one by one based on data from both the target and auxiliary domains. The second step is testing, in which, based on the detected auxiliary domains, independence tests are conducted and a majority voting scheme is then used to aggregate their information; the second step no longer involves the target data. To avoid potential error accumulation due to the dependence of the two steps, we adopt a sample-splitting strategy for the auxiliary domains. Specifically, for each auxiliary domain $k$, the sample is equally split into $\mathbf{X}^{(k,1)}$ and $\mathbf{X}^{(k,2)}$, which are used to perform the two key steps of detection and testing, respectively.}

   In the first step of detection, to capture the local similarity with respect to node $j$, we introduce a novel index to detect the informative auxiliary DAGs. Specifically, for any node $j \in \widehat{\cal S}_t$, we define
   \begin{equation}\label{k_node}
   	\begin{aligned}
   		& \widehat{\xi}_{j,t}^{(k)} = \Big | \max_{l \in \widehat{\cal S}_t \setminus \{j\}} \widehat{\mbox{dcov}}^2 (\widehat{\be}_{j,\widehat{\cal S}_t}, {\bx}_l) - {\max_{l^{\prime} \in \widehat{\cal S}_t \setminus \{j\}} \widehat{\mbox{dcov}}^2 (\widehat{\be}_{j,\widehat{\cal S}_t}^{(k,1)}, {\bx}_{l^{\prime}}^{(k,1)}) } \Big |,
   	\end{aligned}
   \end{equation}
   {where $\max_{l^{\prime} \in \widehat{\cal S}_t \setminus \{j\}} \widehat{\mbox{dcov}}^2 (\widehat{\be}_{j,\widehat{\cal S}_t}^{(k,1)}, {\bx}_{l^{\prime}}^{(k,1)})$ is calculated using the first part of the $k$-th auxiliary domain.} This measures the difference in the maximum distance covariance between the $j$-th expected residual and other variables in $\widehat{\cal S}_t$ between the $k$-th auxiliary domain and the target domain. 
   The selection criterion in \eqref{k_node} is motivated by some key observations at the population level. On one hand, we have $\max_{l \in {\cal S}_t \setminus \{j\}} \mbox{dcov}^2 (e_{j,{\cal S}_t}, x_l) = 0$ for any node $j \in {\cal A}_{t}$. If ${\cal G}_k$ is node-$j$ structure-informative, then $\max_{l^{\prime} \in {\cal S}_t \setminus \{j\}} \mbox{dcov}^2 (e_{j,{\cal S}_t}^{(k)}, x_{l^{\prime}}^{(k)}) = 0$. Yet, for any other non-informative ${\cal G}_{\tilde{k}}$ that does not satisfy Definition \ref{sim-struc}c(i), $\max_{l^{\prime} \in {\cal S}_t \setminus \{j\}} \mbox{dcov}^2 (e_{j,{\cal S}_t}^{(\tilde{k})}, x_{l^{\prime}}^{(\tilde{k})})$ shall be substantially away from 0. Therefore, the node-$j$ structure-informative under Definition \ref{sim-struc}c(i) must have smaller $\widehat{\xi}_{j,t}^{(k)}$ value. On the other hand, $\max_{l \in {\cal S}_t \setminus \{j\}} \mbox{dcov}^2 (e_{j,{\cal S}_t}, x_l)$ is substantially away from 0 for any node $j \in {\cal S}_t \setminus {\cal A}_{t}$. Therefore, any non-informative ${\cal G}_k$ with $\max_{l^{\prime} \in {\cal S}_t \setminus \{j\}} \mbox{dcov}^2 (e_{j,{\cal S}_t}^{(k)}, x_{l^{\prime}}^{(k)}) = 0$ must have larger $\widehat{\xi}_{j,t}^{(k)}$ value.
   A natural way is to select the most informative auxiliary DAG as $k_{j,t} = \argmin_{k \in [K]} \widehat{\xi}_{j,t}^{(k)}$, but it may not fully utilize information from multiple underlying structure-informative auxiliary DAGs. To improve the performance of structural transfer learning, we introduce a thresholding constant $\xi_{t}^*$, then we can declare multiple informative auxiliary DAGs with $\widehat{\xi}_{j,t}^{(k)}$ smaller than $\xi_{t}^*$ for each node $j \in \widehat{\cal S}_t$, whose indices are denoted as
   \begin{equation}\label{k_node2}
   	\begin{aligned}
   		\widehat{\cal K}_{j,t}^{N} = \{ k: \widehat{\xi}_{j,t}^{(k)} <  \xi_{t}^* \},
   	\end{aligned}
   \end{equation}
   and the superscript $N$ represents the \textbf{N}ode-level structure-informative set to distinguish the superscript $P$ used in ${\cal K}_{t}^P$ of Condition \ref{K_para} representing a \textbf{P}arameter-informative set.
   The auxiliary domains included in the set \eqref{k_node2} with an appropriate value of $\xi_{t}^*$ must be useful for assigning node-$j$ when constructing $\mathcal{A}_t$ and also connected with Definition \ref{sim-struc}c.
   
   As for the selection of $\xi_{t}^*$, we define $\rho_{0,t}^2 = \min_{j \in {\cal S}_t \backslash {\cal A}_t} \max_{l \in {\cal S}_{t}\backslash \{j\}} \mbox{dcov}^2 (e_{j,{\cal S}_{t}}, x_l)$ for $t=0,..., T-2$ at the population level and $\rho_{0,t}^2 = C(\max\{ d_t^{3/2} \sqrt{\frac{ \log \gamma_{t,n} }{n} }, n^{-\eta}\})$ for $t=T-1$ and a sufficiently large constant $C$, with $d_t = \| \bfTheta^{{\cal S}_t *} \|_{0, \infty}$, $\gamma_{t,n} = p I(t=0) + \max\{ |{\cal S}_t|, n \} I(t>0)$, and $0<\eta<\frac{1}{2}$.
   By the property of distance covariance, $\rho_{0,t}^2 > 0$ and $\max_{l \in {\cal S}_{t}\backslash \{j\}} \mbox{dcov}^2 (e_{j,{\cal S}_{t}}, x_l) = 0$ for $j \in {\cal A}_t$. 
   Note that the theoretical result established in Section \ref{property} guarantees that the value of $\xi_{t}^*$ can be specified as $\rho_{0,t}^2/c$ with some constant $c>1$. 
   In practice, at step $t$, we first apply the target-only layer-identification procedure to the target observations restricted to the current remaining-node set $\widehat{\cal S}_t$, and denote the resulting preliminary leaf set by $\widehat{\cal A}_t^{(0)}$. When $\widehat{\cal S}_t\backslash\widehat{\cal A}_t^{(0)}\ne\emptyset$, we set $\widehat{\rho}_{0,t}^2 = \min\limits_{j \in \widehat{\cal S}_t \backslash \widehat{\cal A}_t^{(0)}} \max\limits_{l \in \widehat{\cal S}_t \setminus \{j\}} \widehat{\mbox{dcov}}^2 (\widehat{\be}_{j,\widehat{\cal S}_t}, {\bx}_l)$, where the empirical distance covariance is defined in \eqref{dcov-def}. If $\widehat{\cal S}_t\backslash\widehat{\cal A}_t^{(0)} = \emptyset$, we set $\widehat\rho_{0,t}^2=\| \widehat{\bfTheta}^{\widehat{\cal S}_t} \|_{0, \infty}^{3/2} \sqrt{\frac{ \log \widehat{\gamma}_{t,n} }{n} }$ empirically with $\widehat{\gamma}_{t,n} = p I(t=0) + \max\{ |\widehat{\cal S}_t|, n \} I(t>0)$. This treatment is adopted for the empirical tuning procedure, whereas the theoretical results are established using the population threshold. Such a way to detect informative auxiliary domains through thresholding constants in empirical research is also widely used in parameter transfer learning \citep{tian2022transfer}. Moreover, following the theoretical suggestions 
   in Section \ref{property} that  $\widehat{\xi}_{j,t}^{(k)} > (1 - o(1) ) \widehat{\rho}_{0,t}^2 $ for the non-informative auxiliary DAG $k$, and $\widehat{\xi}_{j,t}^{(k)} < (\kappa + o(1) ) \widehat{\rho}_{0,t}^2 $ with some constant $0 < \kappa < 1/2$ for the informative auxiliary DAG $k$, the constant $c$ can be set as $2$.
   In all the numerical examples, we set the empirical value of $\xi_{t}^*$  as $\widehat{\rho}_{0,t}^2/2$, which yields satisfactory performance.

   In the second step of testing, once $\widehat{\cal K}_{j,t}^{N}$ is determined for each node $j$, $\widehat{\cal A}_{t}$ can be obtained by combining the distance covariance-based independence tests between $\widehat{{\be}}_{j,\widehat{\cal S}_t}^{(k,2)}$ and $\{{\bx}_l^{(k,2)}: l \in  \widehat{\cal S}_t\backslash \{j\} \}$ based on the second part of the $k$-th auxiliary sample for $k \in \widehat{\cal K}_{j,t}^{N}$,
   \begin{align}\label{A_node2}
   	\widehat{\cal A}_{t} = 
   	\Big \{ j: \sum_{k \in \widehat{\cal K}_{j,t}^{N}} I \Big[ \widehat{{\be}}_{j,\widehat{\cal S}_t}^{(k,2)} \ \text{is tested to be  independent with} \ {\bx}_l^{(k,2)}, \
   	\text{for any} \ l \in  \widehat{\cal S}_t\backslash \{j\}    \Big] > |\widehat{\cal K}_{j,t}^{N}| / 2 \Big \}, 
   \end{align}
   Each auxiliary domain first forms a node-level decision by requiring all the coordinatewise independence tests to accept independence. These node-level decisions are then combined by strict majority voting; a tied vote excludes the node from the current layer.

   The independence test in \eqref{A_node2} is exclusively based on the detected informative auxiliary DAGs that share exactly the similar local structure as the target DAG, which can outperform single DAG learning \citep{zhao2022learning} with a smaller sample size. In the proposed local structural transfer, the target domain is only used to efficiently detect informative DAGs via the well-designed criterion in \eqref{k_node}, which shares an analogous spirit with bias correction using the target domain only in the parameter transfer procedure \citep{li2022transferb, li2022transfera, tian2022transfer}.
   
   As a high-level summary, the following proposition reveals a direct connection between the proposed detection criteria \eqref{k_node}--\eqref{k_node2} and the definition of node-level structure-informative DAGs in Definition \ref{sim-struc}c, and explains why the discrepancy statistic in \eqref{k_node} can reliably distinguish node-level structure-informative DAGs from non-informative ones. A rigorous formulation is provided in the Supporting Information, and a more detailed discussion of the connection between Definition \ref{sim-struc}c and Proposition \ref{detec_node0-main} is provided in Section A.1.1 of the Supporting Information.
   \begin{prop}\label{detec_node0-main}
   	When constructing ${\cal A}_{t}$ after ${\cal A}_0, \cdots, {\cal A}_{t-1}$ have been constructed, let $\widehat{\cal K}_{j,t}^{N}$ denote the set of auxiliary DAGs detected for node $j$ using \eqref{k_node} and \eqref{k_node2} with $\xi_{t}^* = \rho_{0,t}^2/2$. Suppose that Conditions \ref{onenorm}--\ref{ass::dcov-p} in Section \ref{property} hold. Then, with probability tending to 1, any $k \in \widehat{\cal K}_{j,t}^{N}$ satisfies the following properties:
   	(i) for any $j \in {\cal A}_t$, $\text{de}_j^{(k)} \cap {\cal S}_{t} = \emptyset$; and (ii) for any $j \in {\cal S}_t \backslash {\cal A}_t$, $\text{de}_j^{(k)} \cap {\cal A}_{t^{\prime}} \ne \emptyset$ for some $t^{\prime} \geqslant t$.
   \end{prop}
   The computational complexity of the local structural transfer method is largely determined by the precision matrix estimation and the independence tests in Steps 2a and 2b of Algorithm 1. The precision matrix estimation in Step 2a has computational complexity $O( |\mathcal{S}_t|^3 (K+1) + |\mathcal{S}_t|^2 \sum_{k=0}^{K} n_k )$, where $n_0$ is the sample size of the target domain. The distance covariance-based independence tests in Step 2b have computational complexity $O( |\mathcal{S}_t|^2 \sum_{k=0}^{K} n_k \log n_k )$. Therefore, the overall computational complexity of Algorithm 1 over $T$ layers is $O(\sum_{t=0}^{T-1} ( |\mathcal{S}_t|^3 (K+1) + |\mathcal{S}_t|^2 \sum_{k=0}^{K} n_k \log n_k ))$. Compared with the single-task TL method of \cite{zhao2022learning}, whose computational complexity is $O(\sum_{t=0}^{T-1} ( |\mathcal{S}_t|^3 + |\mathcal{S}_t|^2 n_0 \log n_0 ))$, the proposed method incurs additional computational cost through the use of $K$ auxiliary domains to incorporate multi-source information. Nevertheless, this additional cost remains manageable in our numerical studies, and the proposed method achieves substantial improvements in statistical accuracy and sample complexity.

   \begin{Rem}[Negative transfer]\label{nega}
   	The developed \eqref{A_node2} relies on the existence of structure-informative auxiliary domains. As a byproduct, introducing $\xi_{t}^*$ can naturally safeguard the extreme case with no informative auxiliary domain to prevent the negative transfer from happening. Specifically, if no auxiliary DAG can be declared, that is, $\widehat{\cal K}_{j,t}^{N}$ is empty, we simply perform the independence test of node $j$ based only on the target domain when constructing $\widehat{\cal A}_{t}$.
   \end{Rem}

   \subsection{Additional parameter transfer}\label{mpt}
   The aforementioned subsections focus on structure transfer without relying on parameter-informative auxiliary DAGs. If one wants to utilize parameter-informative auxiliary DAGs to enhance the estimation of edge strength, in Step 2a of Algorithm \ref{alg:2}, the well-established parameter transfer learning on the target precision matrix \citep{li2022transfera} or its variant can be employed, and more detailed discussions are provided in the Supporting Information.

   \subsection{Theoretical properties}\label{property}
   
   This section presents some theoretical properties of the proposed local structural transfer learning method for DAGs in two cases without and with parameter-informative DAGs, including the exact DAG reconstruction consistency of $\widehat{\cal G}_0$ and $\widehat{\bB}$.
   Define $\widetilde{n}=\min_{k\in[K]}n_k$,
   $d_t=\|\bfTheta^{{\cal S}_t*}\|_{0,\infty}$,
   $\gamma_{t,n}=pI(t=0)+\max\{|{\cal S}_t|,n\}I(t>0)$,
   and $\delta_{h_t}=(1+h_t)(h_t\sqrt{\log\gamma_{t,n}/n}\wedge h_t^2)$.
   For auxiliary-domain calculations at step $t$, define
   $d_t^{(k)}=\|[\bfTheta^{{\cal S}_t}]^{(k)*}\|_{0,\infty}$
   and $\gamma_{t,n_k}^{(k)}
   =pI(t=0)+\max\{|{\cal S}_t|,n_k\}I(t>0)$,
   where $[\bfTheta^{{\cal S}_t}]^{(k)*}$ is the population precision
   matrix of $\bx_{{\cal S}_t}^{(k)}$.
   Define $\tilde{\varrho}_{k,t}^{2}
   =C\max\{(d_t^{(k)})^{3}
   \sqrt{\log\gamma_{t,n_k}^{(k)}/n_k},n_k^{-\eta}\}$
   for $k\in[K]$ and $t=0,\ldots,T-1$, where $0<\eta<1/2$
   and $C$ is a sufficiently large constant.
   Throughout this section, suppose that
   $\log p=O(n^{(1-2\eta)/3})$ and
   $K\leqslant C\min\{p,n\}^{C}$.
   
   \begin{Con}\label{onenorm}
   	$\max_{0\leqslant t<T}
   	\max\{\|\bfTheta^{{\cal S}_t*}\|_{1,\infty},
   	\max_{k\in[K]}\|[\bfTheta^{{\cal S}_t}]^{(k)*}\|_{1,\infty}\}$
   	is bounded, and there exists a constant $C_1$ such that
   	$1/C_1\leqslant\psi_{\min}(\bfSigma^*)
   	\leqslant\psi_{\max}(\bfSigma^*)\leqslant C_1$
   	and
   	$1/C_1\leqslant\psi_{\min}(\bfSigma^{(k)*})
   	\leqslant\psi_{\max}(\bfSigma^{(k)*})\leqslant C_1$
   	for any $k\in[K]$.
   	For any $j\in{\cal N}$ and $k\in[K]$,
   	both $\epsilon_j/\sigma_j$ and
   	$\epsilon_j^{(k)}/\sigma_j^{(k)}$ are sub-Gaussian,
   	with uniformly bounded sub-Gaussian norms.
   \end{Con}

   \begin{Con}\label{Irrep}
   	Denote ${\cal C}_{tj}=\{ l \in {\cal S}_t : [\bfTheta^{{\cal S}_t *}]_{jl} \neq 0\}$, ${\cal C}_{tj}^c = {\cal S}_t \backslash {\cal C}_{tj}$, and $\bfSigma^{*}_{{\cal C}_{tj} {\cal C}_{tj}}$ the sub-matrix with rows and columns of $\bfSigma^{*}$ indexed by ${\cal C}_{tj}$ and ${\cal C}_{tj}$, respectively. 
   	There exists some constant $C_2 \in (0,1]$ such that $\max_{t\in \{0,...,T-1\}} \max_{j \in {\cal S}_t, e \in {\cal C}_{tj}^c} \| \bfSigma^{*}_{e {\cal C}_{tj}} (\bfSigma^{*}_{{\cal C}_{tj} {\cal C}_{tj}})^{-1} \|_1 \leqslant 1-C_2$ and $\max_{k \in [K]} \max_{t\in \{0,...,T-1\}} \max_{j \in {\cal S}_t, e \in {\cal C}_{tj}^c} \| \bfSigma^{(k)*}_{e {\cal C}_{tj}} (\bfSigma^{(k)*}_{{\cal C}_{tj} {\cal C}_{tj}})^{-1} \|_1 \leqslant 1-C_2$.
   \end{Con}
   
   
   %
   %

   \begin{Con}\label{ass::dcov-glo}
   	Let $T^{(k)}$ denote the number of layers in ${\cal G}_k$. For any $k \in [K]$, a sufficiently large $C$, and $t =0, 1, \cdots, T^{(k)}-2$, $\min_{j \in {\cal S}_t^{(k)} \backslash {\cal A}_t^{(k)}} \max_{l \in {\cal S}_{t}^{(k)} \backslash \{j\}} \mbox{dcov}^2 (e^{(k)}_{j,{\cal S}_{t}^{(k)}}, x_l^{(k)}) \geqslant C \max \{ (d_t^{(k)})^{3} \sqrt{n_k^{-1} \log \gamma_{t,n_k}^{(k)}}, n_k^{-\eta}\}$. For any $k\in[K]$ and $t=0,\ldots,T-1$,
   	$\tilde{\varrho}_{k,t}^{2}\leqslant\rho_{0,t}^{2}/8$.
   	Meanwhile, for any $j\in{\cal S}_t$,
   	$\max_{l\in{\cal S}_t\backslash\{j\}}
   	\mbox{dcov}^{2}(e_{j,{\cal S}_t}^{(k)},x_l^{(k)})=0$
   	if and only if
   	$\text{de}_j^{(k)}\cap{\cal S}_t=\emptyset$.
   \end{Con}
   
   \begin{Con}\label{K_node_diff_weak}
   	For any $j\in{\cal S}_0$, if
   	$\tilde{k}\notin{\cal K}_j^N$, then
   	$\max_{l^{\prime}\in{\cal S}_{t_j}\backslash\{j\}}
   	\mbox{dcov}^2(e_{j,{\cal S}_{t_j}}^{(\tilde{k})},
   	x_{l^{\prime}}^{(\tilde{k})})$
   	is either zero or at least $\rho_{0,t_j}^2$; and for $j\in{\cal S}_0\setminus{\cal A}_0$, if
   	$k\in{\cal K}_j^N$,
   	$|\max_{l\in{\cal S}_t\backslash\{j\}}
   	\mbox{dcov}^2(e_{j,{\cal S}_t},x_l)
   	-\max_{l\in{\cal S}_t\backslash\{j\}}
   	\mbox{dcov}^2(e_{j,{\cal S}_t}^{(k)},x_l^{(k)})|
   	<\kappa\rho_{0,t}^2$
   	for any $0\leqslant t<t_j$ and some constant
   	$0<\kappa<1/2$.
   \end{Con}
   
   Condition \ref{onenorm} prevents both $\bfTheta^*$ and $\bfTheta^{(k)*}$ from divergence and characterizes the noise distribution implying that ${x_j}/{\sqrt{\Sigma_{jj}}}$ also follows a sub-Gaussian distribution, which has been commonly employed in literature \citep{lam2009sparsistency,zhao2022learning}.
   Condition \ref{Irrep} limits the correlation between the zero and non-zero elements in both $\bfSigma^*$ and $\bfSigma^{(k)*}$. This is analogous to the popular irrepresentable condition required for many lasso-type methods \citep{ravikumar2011high,zhang2014sparse,liu2015fast}. Condition \ref{ass::dcov-glo} ensures that the different topological layers are sufficiently discriminated from the distance covariance-based correlation measure on auxiliary domains. In fact, for nodes in ${\cal S}_t^{(k)} \backslash {\cal A}_t^{(k)}$, it is required that $\min_{j \in {\cal S}_t^{(k)} \backslash {\cal A}_t^{(k)}} \max_{l \in {\cal S}_{t}^{(k)} \backslash \{j\}} \mbox{dcov}^2 (e^{(k)}_{j,{\cal S}_{t}^{(k)}}, x_l^{(k)})$ is away from 0 by a certain order, which can be regarded as the minimum signal of distance covariance between different layers. $0<\eta<\frac{1}{2}$ is a constant that controls the order away from 0, which is common in existing literature \citep{zhao2022learning}. 
   Condition \ref{K_node_diff_weak}, as the detectable condition of structure-informative auxiliary domains, assures that informative auxiliary domains shall be sufficiently closer to the target domain than the non-informative auxiliary domain, so that the informative auxiliary domains can be efficiently detected.
   Such detectable conditions have been widely used in the literature of transfer learning, although their actual form may vary under different models \citep{tian2022transfer}. To verify Condition \ref{K_node_diff_weak}, encompassing a wide range of auxiliary DAG scenarios, is not an impractical endeavor. Indeed, a variety of intuitive examples are constructed in numerical simulations.

   \subsubsection{Local structural transfer without parameter-informative DAG}
   We now establish the asymptotic exact DAG recovery via local transfer in the case with unknown node-level structure-informative auxiliary DAGs and without parameter-informative DAGs. Some additional conditions are necessary.
   
   \begin{Con}\label{ass::dcov-p} 
   	For any $t=0,..., T-2$, $\rho_{0,t}^2 = \min_{j \in {\cal S}_t \backslash {\cal A}_t} \max_{l \in {\cal S}_{t}\backslash \{j\}} \mbox{dcov}^2 (e_{j,{\cal S}_{t}}, x_l) \geqslant C(\max\{ d_t^{3/2} \sqrt{\frac{ \log \gamma_{t,n} }{n} }, n^{-\eta}\})$ for a sufficiently large constant $C$; $\max_{l \in {\cal S}_{t}\backslash \{j\}} \mbox{dcov}^2 (e_{j,{\cal S}_{t}}, x_l) = 0$ for $j \in {\cal A}_t$, $t=0,..., T-1$.
   \end{Con}
   
   Condition \ref{ass::dcov-p} prevents the distance covariance from being too small in the target DAG to discriminate nodes in ${\cal A}_t$ and ${\cal S}_t \backslash {\cal A}_t$. More interestingly, the term $d_t^{3/2} \sqrt{\frac{ \log \gamma_{t,n} }{n} }$ can be substantially smaller than $d_t^{3} \sqrt{\frac{ \log \gamma_{t,n} }{n} } $ as in the single-task learning for DAG \citep{zhao2022learning}. Then we first give a consistency of node-level assignment.

   \begin{Mth}[{\bf Node-wise assignment consistency}]
   	\label{thm::consis::At-main-p-j}
   	When constructing ${\cal A}_t$ using \eqref{A_node2}, set the significance level of the independence test as
   	$2\{1-\Phi(C_{\alpha}\sqrt{n_{k2}} \tilde{\varrho}_{k,t })\}$.
   	Suppose Conditions \ref{onenorm}--\ref{ass::dcov-p} hold and $\xi_t^*=\rho_{0,t}^2/2$. Then there exist positive constants $c$, $C_\alpha$, and $\tau>2$ such that, for node $j\in{\cal A}_0$,
   	\[
   	\begin{aligned}
   		\pr(j\notin\widehat{\cal A}_0 )
   		\leqslant{}&
   		cp^{1-\tau}
   		+c\exp\{-cn^{(1-2\eta)/3}\}
   		+c\big[p\zeta_j(n,\widetilde n)\big]^{k_j^*},
   	\end{aligned}
   	\]
   	where
   	$\zeta_j(n,\widetilde n)
   	=4\exp\{-cn^{(1-2\eta)/3}\}I(|{\cal K}_j^N|=0)
   	+4\exp\{-c\widetilde n^{(1-2\eta)/3}\}I(|{\cal K}_j^N|>0)$,
   	$\widetilde n=\min_{k\in[K]}n_k$, and
   	$k_j^*=\max\{1,\lceil|{\cal K}_j^N|/2\rceil\}$.
   	Here, ${\cal K}_j^N$ is the node-$j$ structure-informative set, and $\eta$ are defined in Condition \ref{ass::dcov-glo}.
   	Furthermore, for node $j\in{\cal A}_t$ with $t>0$, 
   	\[
   	\begin{aligned}
   		&  	\pr (j \notin \widehat{\cal A}_t, j \in {\cal A}_t, \widehat{\cal A}_{0}  = {\cal A}_0, \cdots, \widehat{\cal A}_{t-1}  = {\cal A}_{t-1}) \\
   		&  \quad \leqslant
   		c(\max\{|{\cal S}_t|,n\})^{1-\tau}
   		+ c\exp\{-cn^{(1-2\eta)/3}\} +
   		c\big[|{\cal S}_t|\zeta_j(n,\widetilde n)\big]^{k_j^*}.
   	\end{aligned}
   	\]
   \end{Mth}

   Theorem \ref{thm::consis::At-main-p-j} establishes the node-wise assignment consistency and ensures each node can be exactly assigned to its correct topological layer with high probability. The first two terms in the tail probability, $c p^{1-\tau} + c \exp\{-c n^{(1-2\eta)/3}\}$, stem from the precision matrix estimation, which is the same as in single-task learning.
   The third term, $c[p\zeta_j(n,\widetilde n)]^{k_j^*}$,
   arises from the node-level independence tests and their aggregation across auxiliary domains. The factor $p$ inside the power accounts for the coordinatewise tests within each auxiliary domain.
   For nodes without informative auxiliary DAGs $(|{\cal K}_j^N|=0)$, $k_j^*=1$ and this term retains the target-domain scale. For nodes with informative auxiliary DAGs $(|{\cal K}_j^N|>0)$, larger auxiliary samples and a larger $k_j^*$ can reduce this term. This reveals the benefit of using multiple informative domains by \eqref{k_node2} rather than only one in the structural transfer step. Under strict majority voting, tied votes exclude the node, so the common exponent for the two types of errors is $k_j^*=\max\{1,\lceil|{\cal K}_j^N|/2\rceil\}$.

   \begin{Mth}[\bf Layer recovery consistency]\label{thm::consis::At-p-main}
   	Under the conditions and notations in Theorem \ref{thm::consis::At-main-p-j}, then there exist some positive constants $c$, $C_{\alpha}$ and $\tau > 2$ such that 
   	\begin{equation*}
   		\begin{aligned}
   			& \pr (\widehat{\cal A}_t  = {\cal A}_t, \text{ for any } t = 0, \cdots, T-1) \\
   			& \geqslant 1-  c p^{2-\tau} - c T n^{2-\tau} - c T p \exp\{-c n^{(1-2\eta)/3}\} - c\sum_{t=0}^{T-1}\sum_{j\in{\cal S}_t}
   			\big[|{\cal S}_t|\zeta_j(n,\widetilde n)\big]^{k_j^*}.
   		\end{aligned}
   	\end{equation*} 
   \end{Mth}
   
   Theorem \ref{thm::consis::At-p-main} ensures that all layers in ${\cal G}_0$ can be exactly recovered by the proposed transfer learning method with high probability. The fourth term of the tail probability, $c\sum_{t=0}^{T-1}\sum_{j\in{\cal S}_t}
   [|{\cal S}_t|\zeta_j(n,\widetilde n)]^{k_j^*}$, reveals the global layer-recovery guarantee by aggregating the node-wise assignment consistency in Theorem \ref{thm::consis::At-main-p-j}. Note that the term $\zeta_{j}(n, \widetilde{n}) = 4 \exp\{-c n^{(1-2\eta)/3}\} I(| {\cal K}^{N}_{j} |=0) +  4 \exp\{-c \widetilde{n}^{(1-2\eta)/3}\} I(| {\cal K}^{N}_{j} |>0) $ indicates that it does not require structure-informative auxiliary DAGs for all nodes. Compared to the corresponding probability term $c p^2 \exp\{-c n^{(1-2\eta)/3}\}$ in the single-task learning \citep{zhao2022learning}, this tail probability can be partially reduced even with only a subset of nodes possessing informative auxiliary DAGs. 
   
   It is also interesting to note that the learning result of topological layer structure depends on $\widetilde{n}$ and $|\mathcal{K}_j^N |$ rather than the quality of auxiliary domains, which is different from the fact that the result of parameter transfer learning depends on the distance between auxiliary and target parameters. To some extent, this reflects the essential difference between structural learning and parameter learning, in which the measure of structural learning effect is not as ``continuous" as the error order in parameter learning. Moreover, the informative DAG detection step in the proposed structure transfer learning framework can exclude the non-informative auxiliary DAGs with probability tending to 1, so that the effect of structure learning only depends on $\widetilde{n}$ and $ |\mathcal{K}_j^N |$, which are related to the informative DAGs, rather than the quality of auxiliary domains.
   
   \begin{Mth}[\bf Graph recovery and estimation consistency]\label{thm:2-main-p}
   	Under conditions and notations in Theorem \ref{thm::consis::At-p-main}, $n \geqslant C \big( (T^{1/(\tau-2)} + d_0^3) [\log (\max\{p,n\})]^{3/(1-2\eta)} \big)$ with $d_0 = \| \bfTheta^{*} \|_{0, \infty}$, and $ \sqrt{\frac{ \log p}{n}} \lesssim  \min_{(j,l) \in \{ (j, l) : \beta^*_{jl} \neq 0\} } | \beta^*_{jl} |$. It then follows that
   	$$\pr (\widehat{\cal G}_0={\cal G}_0) \rightarrow 1 \ \mbox{as} \ n \rightarrow \infty, \ p \rightarrow \infty, $$ 
   	and $\|\widehat{\bB} -  \bB^* \|_{F} = O_p \left( \sqrt{ |{\cal E}_0| \frac{ \log p}{n} } \right)$. 
   \end{Mth}
   
   Theorem \ref{thm:2-main-p} establishes the asymptotic exact recovery of ${\cal G}_0$ and parameter estimation consistency for the linear non-Gaussian DAG, which holds even for the high dimensional case with $p = O \big( \exp (n^{c_{\eta}}) \big)$ with $c_{\eta} = \frac{1-2\eta}{3}$ and $0<\eta<\frac{1}{2}$, if $T$ and $d_0$ are constants.
   It is also important to remark that the proposed local transfer method by exploiting the node-level structure-informative auxiliary DAGs relaxes the sample complexity from $n \geqslant C \big(T^{1/(\tau-4)} d_0^6 [\log (\max\{p,n\})]^{3/(1-2\eta)}\big)$ \citep{zhao2022learning} to $n \geqslant C \big( (T^{1/(\tau-2)} + d_0^3) [\log (\max\{p,n\})]^{3/(1-2\eta)} \big)$. Note that for such local structural transfer without parameter-informative DAGs, the estimation convergence rate and the minimum signal of $\bB^*$ can not be improved compared to those using the target domain only.
   
   \subsubsection{Local structural transfer with parameter-informative DAG}
   If one performs parameter transfer to utilize parameter-informative auxiliary DAGs to enhance the estimation of edge strength, the asymptotic exact DAG recovery can be further improved. 
   
   \begin{Con}\label{K_para}
   	The parameter-informative set ${\cal K}_{t}^P = \{ k: \|  \bfDelta^{(k)*}_{{\cal S}_t} \|_{1, \infty} + \|  ( \bfDelta^{(k)*}_{{\cal S}_t} )^{\top} \|_{1, \infty} \leqslant h_t \}$ is non-empty for any $ t = 0,1, \cdots, T-1$ and $h_t \lesssim d_t \sqrt{\frac{\log \gamma_{t,n}}{n}} \lesssim 1$. Moreover,
   	$\min_{\tilde{k} \notin {\cal K}_{t}^P} \|  \bfDelta^{(\tilde{k})*}_{{\cal S}_t} \|_{1, \infty} \gg d_t \sqrt{\frac{\log \gamma_{t,n} }{n}}$.
   \end{Con}
   
   \begin{Con}\label{ass::dcov} 
   	For any $t=0,..., T-2$, $\rho_{0,t}^2 = \min_{j \in {\cal S}_t \backslash {\cal A}_t} \max_{l \in {\cal S}_{t}\backslash \{j\}} \mbox{dcov}^2 (e_{j,{\cal S}_{t}}, x_l) \geqslant C(\max\{ d_t^{3/2} (\sqrt{\frac{\delta_{h_t}}{d_t}} + \sqrt{\frac{ \log \gamma_{t,\widetilde{n}} }{\widetilde{n}+n}}), n^{-\eta}\})$ with $0<\eta<\frac{1}{2}$ and a sufficiently large constant $C$. And $\max_{l \in {\cal S}_{t}\backslash \{j\}} \mbox{dcov}^2 (e_{j,{\cal S}_{t}}, x_l) = 0$ for $j \in {\cal A}_t$ and $t=0,..., T-1$.
   \end{Con}

   Similar to Condition \ref{K_node_diff_weak}, Condition \ref{K_para} assures that there exists at least one parameter-informative auxiliary domain for each topological layer, and the second half of the condition can be regarded as the detectable condition of the informative auxiliary domains. 
   More interestingly, benefiting from parameter transfer, Condition \ref{ass::dcov-p}, preventing the distance covariance from being too small in the target DAG to discriminate nodes in ${\cal A}_t$ and ${\cal S}_t \backslash {\cal A}_t$, can be relaxed to Condition \ref{ass::dcov}. Specifically, when $\widetilde{n} \gg n$ and $h_t \ll d_t \sqrt{\frac{\log \gamma_{t,n}}{n}}$, the term $d_t^{3/2} (\sqrt{\frac{\delta_{h_t}}{d_t}} + \sqrt{\frac{ \log \gamma_{t,\widetilde{n}}}{\widetilde{n}+n}})$ in Condition \ref{ass::dcov} can be substantially smaller than $d_t^{3/2} \sqrt{\frac{ \log \gamma_{t,n} }{n} }$ in Condition \ref{ass::dcov-p}. 
   
   The following two theorems demonstrate the theoretical improvement of the local structural transfer in the case where there are parameter-informative auxiliary DAGs.
   
   \begin{Mth}[\bf Layer recovery consistency]\label{thm::consis::At-main}
   	Suppose Conditions \ref{onenorm} - \ref{K_node_diff_weak} and Conditions \ref{K_para} - \ref{ass::dcov} hold, and $\xi_{t}^* = \rho_{0,t}^2/2$, then there exist some positive constants $c$, $C_{\alpha}$ and $\tau > 2$ such that 
   	\begin{equation*}
   		\begin{aligned}
   			& \pr (\widehat{\cal A}_t  = {\cal A}_t, \text{ for any } t = 0, \cdots, T-1) \\
   			& \geqslant 1-  c p^{2-\tau} - c T n^{2-\tau} - c T p \exp\{-c n^{(1-2\eta)/3}\} - {c \sum_{t=0}^{T-1} \sum_{j \in {\cal S}_t} \big[|{\cal S}_t|\zeta_j(n,\widetilde n)\big]^{k_j^*}}.
   		\end{aligned}
   	\end{equation*} 
   \end{Mth}
   
   \begin{Mth}[\bf Graph recovery and estimation consistency]\label{thm:2-main}
   	Suppose that all the conditions in Theorem \ref{thm::consis::At-main} hold, $n \geqslant C \big( (T^{1/(\tau-2)} + d_0^3) [\log (\max\{p,n\})]^{3/(1-2\eta)} \big)$, and $ \sqrt{\frac{\delta_{h_0}}{d_0}} + \sqrt{\frac{ \log p}{\widetilde{n}+n}} \lesssim  \min_{(j,l) \in \{ (j, l) : \beta^*_{jl} \neq 0\} } | \beta^*_{jl} |$. It then follows that
   	$$
   	\pr (\widehat{\cal G}_0={\cal G}_0) \rightarrow 1 \ \mbox{as} \ n \rightarrow \infty, \ p \rightarrow \infty,
   	$$
   	and $\|\widehat{\bB} -  \bB^* \|_{F} = O_p \left( \sqrt{ |{\cal E}_0| \frac{\delta_{h_0}}{d_0} + |{\cal E}_0| \frac{ \log p}{\widetilde{n}+n} } \right)$. 
   \end{Mth}
   
   It is interesting to note that, when $\widetilde{n} \gg n$ and $h_0 \ll d_0 \sqrt{\frac{\log p}{n}}$, we have $\|\widehat{\bB} -  \bB^* \|_{F} = O_p \left( \sqrt{ |{\cal E}_0| \frac{\delta_{h_0}}{d_0} + |{\cal E}_0| \frac{ \log p}{\widetilde{n}+n} } \right)$, which achieves a faster convergence rate than $O_p \left( \sqrt{ |{\cal E}_0| \frac{ \log p}{n} } \right)$ in Theorem \ref{thm:2-main-p}. Moreover, the minimum signal of $\bB^*$ can be relaxed to $ \sqrt{\frac{\delta_{h_0}}{d_0}} + \sqrt{\frac{ \log p}{\widetilde{n}+n}}$ from $\sqrt{ \frac{ \log p}{n} }$ in Theorem \ref{thm:2-main-p}. These show the further improvement after performing parameter transfer.

   \subsubsection{Theoretical comparisons}
   Table \ref{theo-comp-main} summarizes the theoretical results and required conditions in different scenarios for a comprehensive comparison. Some interesting observations, with a focus on the scenario without a parameter-informative auxiliary DAG, can be made.
   
   \begin{table}[!htb]
   	\caption{Theoretical comparisons of different scenarios, where P stands for parameter-informative, Node stand for node-level structure-informative, and Single stands for single DAG learning with no auxiliary DAGs.}
   	\scalebox{0.6}{
   		\begin{tabular}{cccccc}
   			\hline
   			& minimum signal of $\bB$ & error of $\widehat{\bB}$ & $^{\dagger}$minimum signal of $\rho_{0,0}^2$ & $^{\ddagger}$tail probability of $\widehat{\cal A}_t = {\cal A}_t$ & target sample complexity	\\
   			\hline			
   			P+Node & $ \sqrt{\frac{\delta_{h_0}}{d_0}} + \sqrt{\frac{ \log p}{\widetilde{n}+n}}$ & $\sqrt{ |{\cal E}_0| ( \frac{\delta_{h_0}}{d_0} + \frac{ \log p}{\widetilde{n}+n}) }$ & $d_0^{3/2} (\sqrt{\frac{\delta_{h_0}}{d_0}} + \sqrt{\frac{ \log p }{\widetilde{n}+n}})$ & $c p^{2-\tau} + c T n^{2-\tau} + c T p \exp\{-c n^{c_{\eta}}\} $ & $ (T^{\frac{1}{\tau-2}} + d_0^3) [\log (\max\{p,n\})]^{\frac{1}{c_{\eta}}}$ \\
   			Node & $\sqrt{\frac{ \log p}{n}}$ & $\sqrt{ |{\cal E}_0| \frac{ \log p}{n}} $ & $d_0^{3/2} \sqrt{\frac{ \log p }{n} }$ & $c p^{2-\tau} + c T n^{2-\tau} + c T p \exp\{-c n^{c_{\eta}}\} $ & $ (T^{\frac{1}{\tau-2}} + d_0^3) [\log (\max\{p,n\})]^{\frac{1}{c_{\eta}}}$ \\
   			Single & $\sqrt{\frac{ \log p}{n}}$ & $\sqrt{ |{\cal E}_0| \frac{ \log p}{n}} $ & $d_0^{3} \sqrt{\frac{ \log p }{n} }$ & $c p^{4-\tau} + c T n^{4-\tau} + c T p^2 \exp\{-c n^{c_{\eta}}\}$ & $ T^{\frac{1}{\tau-4}} d_0^6 [\log (\max\{p,n\})]^{\frac{1}{c_{\eta}}}$ \\
   			\hline
   	\end{tabular}}
   	\footnotesize{
   		$^{\dagger}$Involving Conditions \ref{ass::dcov-p} and \ref{ass::dcov}: $\delta_{h_0} = (1 + h_0) (h_0 \sqrt{\frac{ \log p }{n}} \wedge h_0^2)$, the common item $n^{-\eta}$ is omitted, and the comparison focuses on $t=0$ for clearer comparison.
   		$^{\ddagger}$In both ``P+Node" and ``Node", the term $c \sum_{t=0}^{T-1} \sum_{j \in {\cal S}_t} [|{\cal S}_t|\zeta_j(n,\widetilde n)]^{k_j^*}$ is omitted, and $c_{\eta} = \frac{1-2\eta}{3} \in (0, \frac{1}{3})$.
   		``P+Node" and ``Node" correspond to Theorem \ref{thm:2-main} and Theorem \ref{thm:2-main-p}, respectively.}
   	\label{theo-comp-main}
   \end{table}
   
   \begin{itemize}
   	\item The estimation error of $\widehat{\bB}$ and the minimum signal of $\bB^*$ cannot be reduced using the local structure transfer method without parameter-informative auxiliary DAGs. This is rather expected as these improvements rely on the parameter transfer.
   	\item The minimum signal condition on the distance covariance $\rho_{0,0}^2$ can be partially improved even without parameter-informative auxiliary DAGs. This is because both parameter transfer and structure transfer can help relax this condition. We also remark that this minimum signal condition implies $d_0^{3/2} \sqrt{\frac{ \log p }{n} } \lesssim 1$, which is still stronger than that in undirected graphical models, as DAG learning is more difficult and challenging than learning undirected graphs. However, it has been better than the best existing result in the literature of single DAG learning \citep{zhao2022learning}, $d_0^{3} \sqrt{\frac{ \log p }{n} } \lesssim 1$. 
   	\item Most importantly, the tail probability of $\widehat{\cal A}_t = {\cal A}_t$ and the target sample complexity to ensure the consistency of $\widehat{\cal G}_0$ can still be improved benefiting from structure-informative DAGs only, and its improved effectiveness is not affected by the presence of parameter-informative auxiliary DAGs.
   \end{itemize}

   We re-underscore the paramount importance of exact DAG recovery over parameter estimation. Crucially, this work demonstrates that the proposed novel structure transfer framework fundamentally reshapes the theoretical landscape surrounding this principle. Specifically, while concerns about parameter estimation error are commonly absent in standard structure learning paradigms, structure transfer introduces significant theoretical advantages. In fact, diminishing the sample complexity to ensure the exact recovery of the target DAG in the context of structure learning can be likened to improving parameter estimation.
   
   \section{Simulation study}
   
   This section examines the numerical performance of the proposed transfer DAG learning methods on three simulated examples. The target and auxiliary DAG structures in the following simulated examples are illustrated in the Supporting Information. The noise $\epsilon_j$'s are randomly generated from either student $t_9$ distribution with variance $\sigma^2_j = (2+4j)^2/12$ or uniform distribution $[-1-2j/p, 1+2j/p]$ for $j = 1, \cdots, p$, ensuring heteroskedasticity of the generated data. The target sample size and dimension are set as $(n,p)=(100,100)$.
   
   {\bf Example 1.} For the target DAG, we consider a sparse hub graph with $T=2$, ${\cal A}_0=\{4,...,p\}$, and ${\cal A}_1=\{1,2,3\}$. For each node in ${\cal A}_1$, we randomly select $\lceil (\log p)^2 \rceil$ nodes in ${\cal A}_0$ as its children, and the coefficient of each directed edge is uniformly generated from $[-1.5,-0.5]\cup[0.5,1.5]$. 
   Three node-level structure-informative DAGs are generated, each with only one hub node selected from $\{1,2,3\}$. In addition, the parameter-informative auxiliary DAG is also generated based on the target DAG, by adding directed edges from half of the isolated nodes in ${\cal A}_0$ to other nodes with probability 0.5. The coefficients of the added directed edges are also uniformly generated from $[-\mu, \mu]$. Its structure is completely different from that of the target DAG, but their parameters are close. Meanwhile, four non-informative auxiliary DAGs with hub nodes that are not in ${\cal A}_1$ are also generated.
   
   {\bf Example 2.} For the target DAG, we consider a DAG consisting of two equal-sized disjoint scale-free subgraphs. Each subgraph is generated by the Barab\'asi-Albert (BA) model, where we start with a single node, and a node with $2$ directed edges is added to the graph at each step. The coefficient of each directed edge is uniformly generated from $[-1.5,-0.5]\cup [0.5,1.5]$. 
   We consider two node-level structure-informative DAGs, each with the same subgraph as the target DAG, while the other subgraph is replaced by an arbitrary scale-free graph. The parameter-informative auxiliary DAG is generated by randomly adding directed edges in the target DAG from nodes in one subgraph to nodes in the other subgraph with a probability of 0.5, regardless of their topological layers. The coefficients of the added directed edges are uniformly generated from $[-\mu, \mu]$. Meanwhile, four non-informative auxiliary DAGs with arbitrary scale-free structures are generated.
   
   In Examples 1 and 2, the number of informative DAGs is fixed, and the sample size of each auxiliary domain is varied as $\widetilde{n} = \omega n$ with $\omega \in \{ 6,10,14 \}$.

   {\bf Example 3.} In this example, we consider block BA graphs similar to those in Example 2 for the target domain and auxiliary DAGs and vary the number of node-level structure-informative auxiliary DAGs. Specifically, there are $\widetilde{\omega}$ pairs of node-level structure-informative auxiliary DAGs with $\widetilde{\omega} \in \{ 0, 2, 4, 6, 8 \}$, and each pair is generated similarly to Example 2. Meanwhile, four non-informative auxiliary DAGs with arbitrary scale-free structures are generated. {Note that $\widetilde{\omega}=0$ corresponds to the case that all the auxiliary domains are non‑informative.}

   We compare our method with some popular linear non-Gaussian single DAG learning methods, such as the topological layer-based learning algorithm (TL; \citealp{zhao2022learning}), the direct high-dimension learning algorithm (MDirect; \citealp{wang2020high}), and the pairwise learning algorithm (Pairwise; \citealp{hyvarinen2013pairwise}); some general single DAG learning methods, including the high dimensional constraint-based PC algorithm (PC; \citealp{kalisch2007estimating}), and a hybrid version of max-min hill-climbing algorithm (MMHC; \citealp{tsamardinos2006max}).
   The proposed method is denoted as Local-Trans. The significance level of independence tests in our methods, TL, and PC is set as $\alpha=0.01$ following the same treatment as in \cite{kalisch2007estimating}. We note that the proposed method is not sensitive to $\alpha$, as confirmed by the sensitivity analysis results in the Supporting Information.
   
   The performances of all the competing methods are measured by several metrics. The true positive rate (TPR), false discovery rate (FDR), and F1-score measure the accuracy of estimated directed edges. The normalized structural Hamming distance (HM; \citealp{tsamardinos2006max}), measuring the smallest number of edge insertions, deletions, and flips needed to convert the estimated DAG into the true one, and Matthews correlation coefficient (MCC; \citealp{yuan2019constrained}) are employed to evaluate the similarity between the estimated and true DAGs. Moreover, the relative error between $\widehat{\bB}$ and ${\bB^*}$, $\text{re}(\widehat{\bB}) = {\| \widehat{\bB}-{\bB^*}\|_F}/{\|  {\bB^*} \|_F}$, is also reported.
   Note that a good estimation is implied with small values of FDR, HM, and $\text{re}(\widehat{\bB})$, but large values of TPR, F1-score, and MCC.

\begin{figure}[!htb]
   	\centering
   	\subfigure{\includegraphics[scale = 0.27]{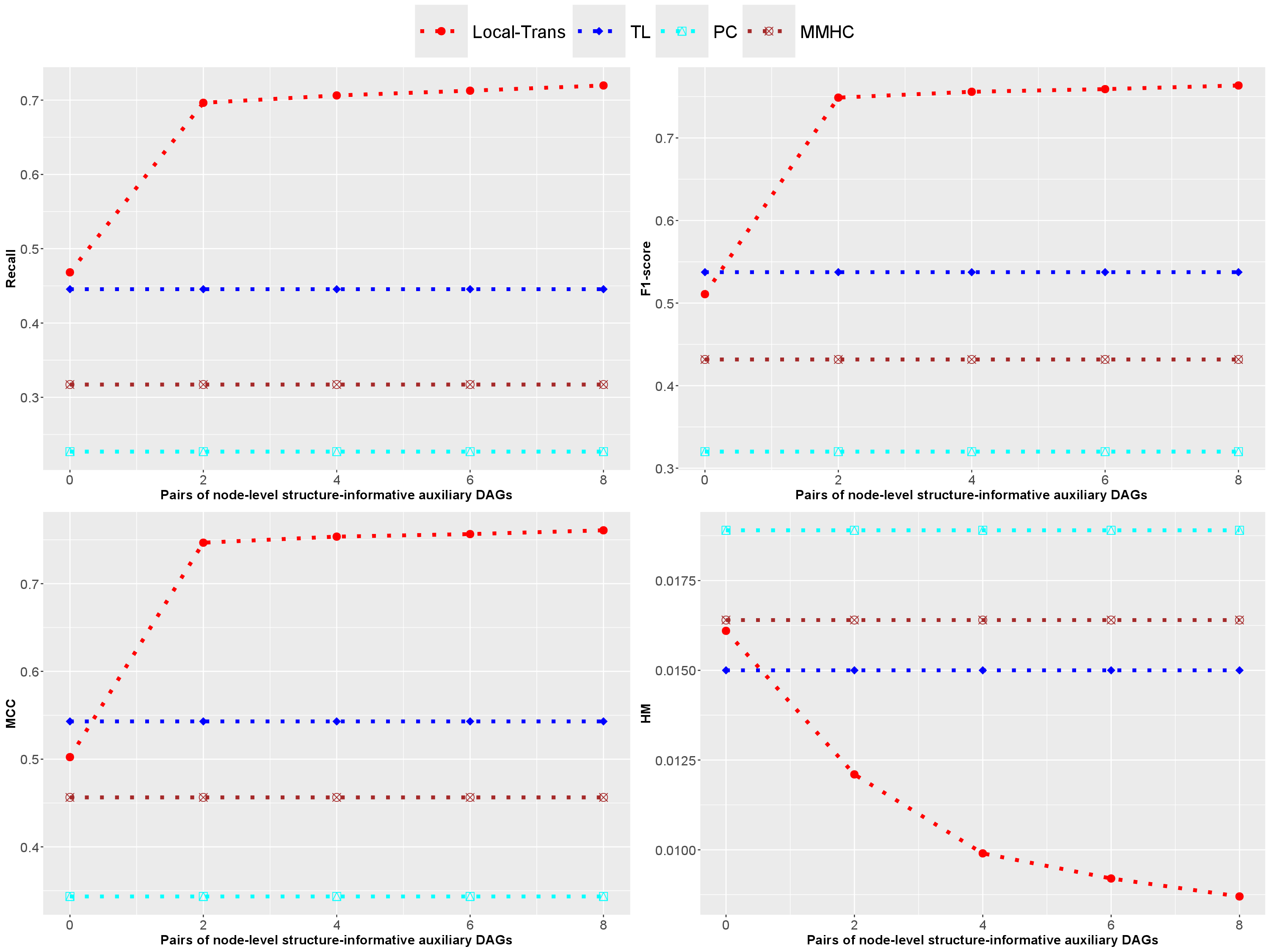}} 
   	\caption{Averaged metrics for Example 3 with student $t$ distributed residuals. Pairwise and MDirect methods are omitted in the figure due to poor performance.}
   	\label{sim-exm3-mvt}
   \end{figure}

   The averaged metrics over 100 independent replications under student-$t$ distributed residuals are summarized in Tables \ref{exa1-mvt} - \ref{exa2-mvt}, and Figure \ref{sim-exm3-mvt}, whereas more numerical results are provided in the Supporting Information. Observations made are very similar under different settings. The proposed Local-Trans method achieves a significant improvement in all metrics compared with its single-task competitors, especially in more challenging Example 2, and the improvement increases as the auxiliary sample size or the number of informative auxiliary DAGs increases. {Moreover, even in the extreme case where no informative auxiliary domain exists ($\widetilde{\omega}=0$), the proposed method demonstrates robust performance in preventing negative transfer.}
   
   We also reported the running time of various methods under all settings in the Supporting Information, and as the number of auxiliary domains and auxiliary sample size increase, compared to single-task learning, the running time of Local-Trans indeed increases accordingly. However, for the analysis of one replicate, Local-Trans only requires less than five minutes under most settings.
   
   \begin{table}[!htb]
   	\centering
   	\caption{Averaged metrics and their standard deviation in parenthesis for Example 1 with student $t$ distributed residuals}
   	\medspace
   	\scalebox{0.8}{
   		\begin{tabular}{cccccccc}
   			\hline
   			\multicolumn{2}{c}{}			& 	Recall	& 	FDR	& 	F1score	& 	MCC	& 	HM	& 	Fnorm	\\
   			\hline
   			\multicolumn{2}{c}{MMHC}			& 	0.0560(0.0277)	& 	0.9697(0.0148)	& 	0.0393(0.0192)	& 	0.0333(0.0203)	& 	0.0166(0.0007)	& 	1.0852(0.0199)	\\
   			\multicolumn{2}{c}{PC}			& 	0.0562(0.0261)	& 	0.9297(0.0316)	& 	0.0623(0.0283)	& 	0.0577(0.0286)	& 	0.0102(0.0005)	& 	1.0259(0.0189)	\\
   			\multicolumn{2}{c}{Pairwise}			& 	0.2220(0.1221)	& 	0.9045(0.0555)	& 	0.1329(0.0752)	& 	0.1374(0.0819)	& 	0.0178(0.0030)	& 	0.9442(0.0498)	\\
   			\multicolumn{2}{c}{MDirect}			& 	0.0317(0.0335)	& 	0.9905(0.0100)	& 	0.0145(0.0153)	& 	0.0065(0.0182)	& 	0.0206(0.0044)	& 	1.5527(0.0031)	\\
   			\multicolumn{2}{c}{TL}			& 	0.9172(0.0412)	& 	0.3233(0.0757)	& 	0.7765(0.0571)	& 	0.7852(0.0527)	& 	0.0033(0.0010)	& 	0.3969(0.0733)	\\
   			Local-Trans		& $\omega=6$	& 	0.8223(0.1958)	& 	0.0797(0.1280)	& 	0.8647(0.1698)	& 	0.8673(0.1656)	& 	0.0014(0.0016)	& 	0.3630(0.2306)	\\
   			& $\omega=10$	& 	0.9353(0.0907)	& 	0.0212(0.0527)	& 	0.9556(0.0738)	& 	0.9561(0.0722)	& 	0.0005(0.0008)	& 	0.2118(0.1390)	\\
   			& $\omega=14$	& 	0.9435(0.0744)	& 	0.0133(0.0451)	& 	0.9640(0.0593)	& 	0.9644(0.0587)	& 	0.0004(0.0006)	& 	0.1935(0.1283)	\\
   			\hline
   	\end{tabular}}
   	\label{exa1-mvt}
   \end{table}
   
   \begin{table}[!htb]
   	\centering
   	\caption{Averaged metrics and their standard deviation in parenthesis for Example 2 with student $t$ distributed residuals}
   	\medspace
   	\scalebox{0.8}{
   		\begin{tabular}{cccccccc}
   			\hline
   			\multicolumn{2}{c}{}			& 	Recall	& 	FDR	& 	F1score	& 	MCC	& 	HM	& 	Fnorm	\\
   			\hline
   			\multicolumn{2}{c}{MMHC}			& 	0.3200(0.0170)	& 	0.3186(0.0396)	& 	0.4353(0.0223)	& 	0.4601(0.0244)	& 	0.0163(0.0007)	& 	0.8546(0.0295)	\\
   			\multicolumn{2}{c}{PC}			& 	0.2252(0.0234)	& 	0.4627(0.0478)	& 	0.3173(0.0311)	& 	0.3399(0.0333)	& 	0.0190(0.0008)	& 	0.9182(0.0254)	\\
   			\multicolumn{2}{c}{Pairwise}			& 	0.3180(0.0464)	& 	0.8970(0.0185)	& 	0.1554(0.0263)	& 	0.1527(0.0306)	& 	0.0681(0.0051)	& 	0.9208(0.0233)	\\
   			\multicolumn{2}{c}{MDirect}			& 	0.0298(0.0139)	& 	0.9802(0.0098)	& 	0.0232(0.0109)	& 	0.0002(0.0124)	& 	0.0452(0.0156)	& 	1.2116(0.0318)	\\
   			\multicolumn{2}{c}{TL}			& 	0.4480(0.0427)	& 	0.3239(0.0390)	& 	0.5378(0.0370)	& 	0.5427(0.0353)	& 	0.0150(0.0010)	& 	1.1244(0.1022)	\\
   			Local-Trans		& $\omega=6$	& 	0.7031(0.0392)	& 	0.2107(0.0333)	& 	0.7434(0.0339)	& 	0.7401(0.0342)	& 	0.0095(0.0012)	& 	0.6965(0.0640)	\\
   			& $\omega=10$	& 	0.7248(0.0463)	& 	0.1963(0.0319)	& 	0.7619(0.0379)	& 	0.7587(0.0380)	& 	0.0089(0.0013)	& 	0.6585(0.0648)	\\
   			& $\omega=14$	& 	0.7339(0.0462)	& 	0.1891(0.0338)	& 	0.7701(0.0385)	& 	0.7670(0.0386)	& 	0.0086(0.0013)	& 	0.6470(0.0649)	\\
   			\hline
   	\end{tabular}}
   	\label{exa2-mvt}
   \end{table}

   \section{Multi-source cancer gene regulatory network analysis}
   
   \subsection{ScRNA-seq data of heterogeneous T cell subgroups in NSCLC}
   
   Understanding T-cell heterogeneity in NSCLC is biologically critical for deciphering mechanisms of immune evasion, predicting immunotherapy responses, and identifying novel therapeutic targets. However, current scRNA-seq studies in NSCLC T cells predominantly focus on defining subgroups and inferring broad functional states based on mean gene expression profiles. Crucially, the analysis of underlying gene regulatory networks, especially directed networks, governing T cell identity, differentiation, and functional specialization remains significantly underdeveloped. Moving beyond static classifications to elucidate these dynamic regulatory interactions is imperative for gaining deeper mechanistic insights into the functional biology of specific T cell subgroups and unlocking their full therapeutic potential within the NSCLC context.
   
   T cell subgroup identification is well-established \citep{guo2018Global}, and substantial heterogeneity implies that gene regulatory networks likely exhibit local rather than global similarity across subgroups. Given the reported association of the {\it hsa05223} pathway with NSCLC, we prioritize its constituent genes for focused network analysis. This targeted approach enhances analytical reliability under limited sample sizes compared to genome-wide analyses. Utilizing single-cell data downloaded from the Gene Expression Omnibus database (\url{https://www.ncbi.nlm.nih.gov/geo/query/acc.cgi?acc=GSE99254}) and \url{http://lung.cancer-pku.cn}, we analyzed 9,055 T cells and 48 pathway genes following removal of features with missing values or standard deviations below 0.25. These T cells were pre-classified into 16 transcriptomically distinct subgroups, with details of names and sample sizes in the Supporting Information, in which CD4-C5-EOMES, a biologically and therapeutically pivotal subgroup, represents a key activated CD4+ T cell state that bridges developmental trajectories and acts as a transitional hub in CD4+ T cell differentiation, and uniquely expresses the critical immunotherapy target PDCD1. Moreover, the CD4-C5-EOMES subgroup comprises only 161 cells, while auxiliary subpopulations contain substantially more cells. Therefore, CD4-C5-EOMES is used as the target domain and other subgroups as the auxiliary domains. 
   
   \begin{figure}[!htb]
   	\centering
   	\subfigure{\includegraphics[scale = 0.3005]{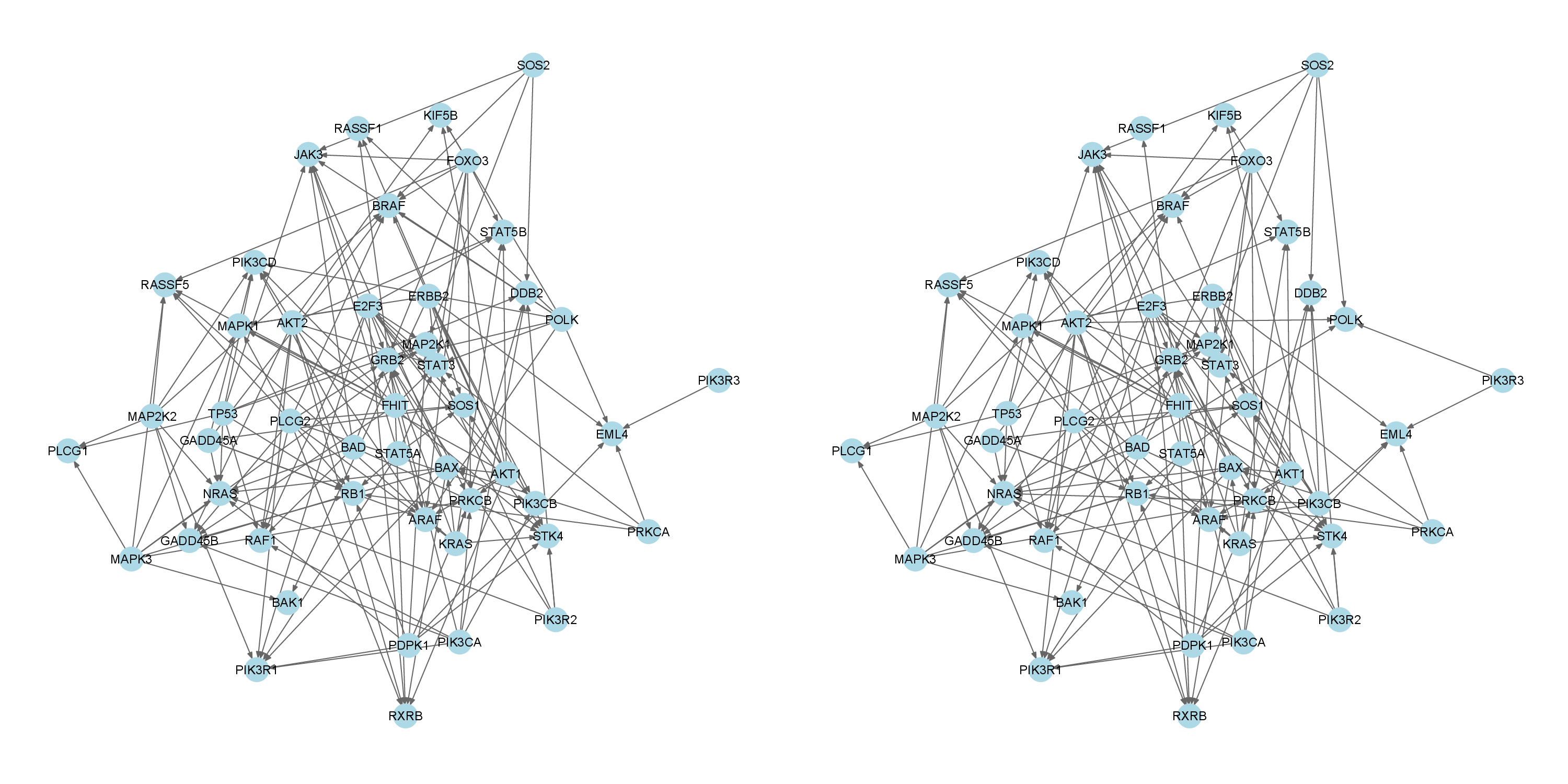}} 
   	\caption{The identified regulatory networks of the CD4-C5-EOMES subgroup using Local-Trans (left) and TL (right). }
   	\label{NSCLC}
   \end{figure}
   
   We applied the proposed local DAG transfer learning method (Local-Trans) to reconstruct its gene regulatory network, overcoming the serious heterogeneity problem. In Figure \ref{NSCLC}, we show the identified network structures. Several hub nodes are important genes reported to be associated with NSCLC.
   For instance, E2F3 is significantly overexpressed in NSCLC tissues compared to adjacent normal lung. This oncogene is directly targeted and suppressed by the tumor suppressor miR-377-3p, forming a critical regulatory axis in NSCLC pathogenesis \citep{sun2016long}. The ERBB2 mutation drives constitutive ligand-independent kinase activation in NSCLC, transforming cells through sustained signaling that sensitizes them to ERBB2 inhibitors, which induce arrest and apoptosis by specifically targeting this mutation, highlighting its therapeutic relevance in this molecular subset. AKT1 critically regulates NSCLC cell survival, proliferation, and migration through isoform-specific mechanisms, distinctively influencing colony formation and motility prominently \citep{lee2011roles}. PLCG2 emerges as an oncogene encoded on extrachromosomal circular DNA, exhibiting upregulated expression in NSCLC that correlates with poor prognosis and drives metastasis by reprogramming mitochondrial respiration dynamics, positioning it as a novel diagnostic and prognostic biomarker \citep{yang2023plcg2}. FOXO3 functions as a tumor suppressor in non-small cell lung cancer, inhibiting malignant cell behaviors. Its tumor-suppressing role positions FOXO3 as a promising therapeutic target for this cancer \citep{zhang2018identification}.

   For comparison, the network structure identified by the single DAG learning method (TL) is also placed. 159 (Local-Trans) and 147 (TL) edges are identified by these two methods. Notably, some hub genes and regulatory edge are only identified by Local-Trans but not by TL. For example, POLK is a hub node regulating many other genes in the network of Local-Trans, but it only receives few edges in the network of TL. In fact, POLK overexpression promotes non-small cell lung cancer development by introducing genetic instability through its error-prone DNA replication, accelerating mutation accumulation and causing DNA damage, chromosomal abnormalities, and increased spontaneous mutagenesis. Specific POLK polymorphisms near its regulatory region demonstrate clinical significance through their association with patient responses to treatment toxicity, suggesting these variants may influence POLK expression levels and consequently drive tumor progression \citep{shao2014association}. In addition, the regulatory relationship from E2F3 to SOS1 is identified by the transfer learning method, but not by the single task learning method. This regulatory relationship has also been reported to be associated with non-small cell lung cancer in existing literature \citep{zhao2018non}.
   
   Moreover, we evaluate the stability of the proposed method and its competitors in detecting DAG structures against sample perturbations, as well as their out-of-sample prediction errors. Specifically, we remove 10\% samples randomly and then estimate the DAG structure with the remaining 90\% samples using Local-Trans and TL, respectively. To evaluate the stability, we calculate the normalized structural Hamming distance (HM) between DAGs estimated previously based on all samples and the remaining 90\% of samples. The process is repeated 100 times, and the average HM of TL and Local-Trans are 0.053 and 0.049, respectively, validating the stability of the proposed method. For prediction errors, similar to the aforementioned process, the remaining 90\% samples are used as the training data in each replicate, yielding coefficient matrix $\widehat{B}_{\text{train}}$, and the removed 10\% samples can be regarded as the test data $\mathbf{X}_{\text{test}}$. The out-of-sample prediction error can be measured by PE = $\frac{1}{\sqrt{n_{\text{test}}}} ||\mathbf{X}_{\text{test}}- \mathbf{X}_{\text{test}} (\widehat{B}_{\text{train}})^T||_F$. The average PE of TL and Local-Trans are 7.14 and 6.96, respectively, demonstrating the satisfactory prediction performance of the proposed method on real data.

   \subsection{Heterogeneous multiple subtypes in breast cancer}

   Breast cancer is molecularly stratified into four primary subtypes defined by receptor expression profiles: Luminal A, Luminal B, HER2-enriched, and triple-negative breast cancer (TNBC). TNBC is specifically identified through immunohistochemical testing that confirms the absence of all three key receptors (ER, PR, and HER2). This subtype exhibits disproportionately high mortality rates, aggressive metastatic behavior, and poor clinical outcomes due to limited targeted therapies, while representing only about 10\% of breast carcinomas. Researching TNBC is critically important to address its therapeutic resistance, uncover novel biomarkers, and develop subtype-specific treatments for this recalcitrant disease. Significant molecular heterogeneity exists in breast cancer, also manifesting in distinct gene regulatory networks across subtypes, influencing therapeutic responses and metastatic pathways. Consequently, deciphering subtype-specific networks provides a mechanistic framework to understand therapeutic vulnerabilities and overcome treatment challenges in TNBC.
   The analyzed data is downloaded from The Cancer Genome Atlas (TCGA) repository (\url{https://portal.gdc.cancer.gov/projects/TCGA-BRCA}), and the {\it hsa05224} pathway is considered, which is related to breast cancer and contains 148 genes.

   Using TNBC as the target domain and Luminal A, Luminal B, HER2-enriched subtypes as auxiliary domains, Local-Trans and TL are applied to identify the gene regulatory network of TNBC, respectively. Due to the numerous nodes and edges, the figure cannot present the network structure clearly; hence tables in the Supporting Information cover the parent and child node information of all edges in the network of Local-Trans and the differential network of the two methods. Some well-known genes related to TNBC, such as KRAS, NOTCH4, CDK4, and DVL1, have been identified as common hub nodes in two networks.
   
   More notably, Local-Trans revealed a unique hub gene absent in TL's output, including genes implicated in TNBC progression, metastasis, or treatment evasion, as evidenced in prior literature. For instance, Akt3 is essential for tumor growth in TNBC, as its depletion potently inhibits growth in multiple cellular models of this disease. Targeting Akt3 alone significantly suppresses TNBC proliferation, demonstrating its pivotal role in driving tumor progression \citep{chin2014targeting}. FGF9 upregulation in cancer-associated fibroblasts, playing a pivotal role in cancer progression, induced cell proliferation, migration, and invasion in vitro, and facilitated tumor growth in vivo in TNBC development \citep{zhang2024upregulation}. HES1 is essential for maintaining cancer stem cell properties in TNBC, whose loss reduces key stemness markers, tumorsphere formation, and clonal growth capacity \citep{li2021hes1}. Abnormalities in the TP53 gene are strongly linked to poor survival outcomes in TNBC patients. The type of TP53 mutation critically determines how its expression level predicts outcome: low expression predicts poor survival with missense mutations, while high expression predicts poor survival with deletion mutations \citep{kim2016association}. FZD6 is frequently overexpressed in TNBC and directly drives cancer cell invasion and metastasis. Reducing FZD6 levels significantly decreases the spread of cancer to distant organs. This pro-metastatic effect occurs through FZD6's role in regulating cell movement pathways, and high levels of FZD6 serve as a strong indicator of increased metastasis risk in TNBC patients \citep{pohl2017wnt}.

   Similar to the previous subsection, we evaluate both stability and prediction performance. TL achieves an average HM of 0.033 and an out-of-sample prediction error of 6.79, while Local-Trans achieves 0.010 and 5.40, respectively. This performance confirms the proposed method's effectiveness on real data regarding both stability and prediction.

   \section{Discussion}
   
   This paper proposes a transfer DAG learning framework, which is able to accommodate informative auxiliary DAGs with different levels of similarity. A novel set of structure similarity measures is also proposed to quantify the global or local similarity between the target and auxiliary DAGs. The substantial theoretical and numerical improvement of the proposed transfer DAG learning in terms of DAG reconstruction in the target study has been demonstrated.  The proposed transfer learning framework may be extended to other structured graphical models, such as non-linear DAG models. 
   
   {It is worth noting that, although multiple testing arises across node pairs, layers, and auxiliary domains, the theoretical guarantees established in Section 4.4 are not formulated in terms of conventional false discovery rate or family-wise error rate control. Instead, our asymptotic consistency results are obtained through a union-of-events argument over the node- and layer-level events, with signal-dependent significance levels that decay to zero at appropriate rates. This ensures exact graph recovery with probability tending to one under the stated conditions, but it does not by itself provide practical FDR or FWER control for a user-specified fixed significance level $\alpha$. Integrating classical procedures such as Benjamini-Hochberg or Bonferroni into the current algorithm framework while preserving theoretical guarantees is not straightforward. We regard the development of principled multiple testing procedures with finite-sample error control in this setting as an important direction for future research.}

   As for the aggregation of multiple informative auxiliary domains $\widehat{\cal K}_{j,t}^{N}$, the majority voting in \eqref{A_node2} does not account for differences in sample size across auxiliary domains. As a heuristic alternative, one may use a weighted aggregation strategy that follows the same order of first summarizing the coordinatewise tests within each auxiliary domain and then aggregating across domains. For each $k \in \widehat{\cal K}_{j,t}^{N}$ and $l \in \widehat{\cal S}_t \setminus \{j\}$, let $\varrho^{(k)}_{j,t,l}$ denote the $p$-value of the independence test between the estimated residual $\widehat{{\be}}_{j,\widehat{\cal S}_t}^{(k,2)}$ and ${\bx}_l^{(k,2)}$. {When $\widehat{\cal K}_{j,t}^{N} \ne \emptyset$, define weights $c_k = n_k / \sum_{k' \in \widehat{\cal K}_{j,t}^{N}} n_{k'}$ and the weighted aggregation score $\bar{\varrho}_{j,t} = \sum_{k \in \widehat{\cal K}_{j,t}^{N}} c_k \min_{l \in \widehat{\cal S}_t \setminus \{j\}} \varrho^{(k)}_{j,t,l}$, with the minimum over an empty coordinate set defined as 1. Under this heuristic rule, node $j$ is assigned to $\widehat{\cal A}_t$ if $\bar{\varrho}_{j,t} > \alpha$, where $\alpha$ is an empirical cutoff. When $\widehat{\cal K}_{j,t}^{N} = \emptyset$, the target-only rule is used. The within-domain minimum reflects the requirement in \eqref{A_node2} that all coordinatewise independence tests within an auxiliary domain must pass before that domain votes to include node $j$. The weighted average, however, is generally not equivalent to majority voting. 
   	It should be emphasized that $\bar{\varrho}_{j,t}$ is an aggregation score rather than a calibrated combined $p$-value, and comparing it with $\alpha$ does not guarantee a test of level $\alpha$. This score-based rule is introduced only as an empirical alternative and is not covered by the theoretical guarantees for \eqref{A_node2}.} A simple numerical comparison in the case of auxiliary domains with different sample sizes is reported in the Supporting Information, and both produced superior performance. Developing more principled weighting strategies and establishing their theoretical properties remain important directions for future research.

   {Moreover, the two tuning parameters in the proposed method, the threshold $\xi_t^*$ for detecting informative auxiliary domains and the significance level $\alpha$ for independence tests, are specified by population-level quantities in theoretical conditions, while in practice they are selected by simple data-driven procedures or empirical choice, such as a rough target-only preliminary estimate and the conventional level $\alpha=0.01$. This discrepancy between theoretical guidance and practical implementation is common in transfer learning.  In the DAG setting, establishing theoretically grounded and practically applicable tuning procedures is particularly challenging because the theoretical orders depend on unknown population quantities and constants that cannot be estimated from data. Developing principled data-adaptive tuning rules with theoretical guarantees remains an important direction for future research in transfer DAG learning.}

\bibliographystyle{jasa3}
\bibliography{reference}

@article{peters2014identifiability,
	title={Identifiability of Gaussian structural equation models with equal error variances},
	author={Peters, Jonas and B{\"u}hlmann, Peter},
	journal={Biometrika},
	volume={101},
	number={1},
	pages={219--228},
	year={2014},
	publisher={Oxford University Press}
}

@article{chen2019causal,
	title={On causal discovery with an equal-variance assumption},
	author={Chen, Wenyu and Drton, Mathias and Wang, Y Samuel},
	journal={Biometrika},
	volume={106},
	number={4},
	pages={973--980},
	year={2019},
	publisher={Oxford University Press}
}

@article{yuan2019constrained,
	title={Constrained likelihood for reconstructing a directed acyclic Gaussian graph},
	author={Yuan, Yiping and Shen, Xiaotong and Pan, Wei and Wang, Zizhuo},
	journal={Biometrika},
	volume={106},
	number={1},
	pages={109--125},
	year={2019},
	publisher={Oxford University Press}
}

@article{shimizu2006linear,
	title={A linear non-Gaussian acyclic model for causal discovery.},
	author={Shimizu, Shohei and Hoyer, Patrik O and Hyv{\"a}rinen, Aapo and Kerminen, Antti and Jordan, Michael},
	journal={Journal of Machine Learning Research},
	volume={7},
	number={10},
pages={2003--2030},
	year={2006}
}

@article{wang2020high,
	title={High-dimensional causal discovery under non-Gaussianity},
	author={Wang, Y Samuel and Drton, Mathias},
	journal={Biometrika},
	volume={107},
	number={1},
	pages={41--59},
	year={2020},
	publisher={Oxford University Press}
}

@article{danks2008integrating,
  title={Integrating locally learned causal structures with overlapping variables},
  author={Danks, David and Glymour, Clark and Tillman, Robert},
  journal={Advances in Neural Information Processing Systems},
  volume={21},
  year={2008}
}

@article{triantafillou2015constraint,
  title={Constraint-based causal discovery from multiple interventions over overlapping variable sets},
  author={Triantafillou, Sofia and Tsamardinos, Ioannis},
  journal={The Journal of Machine Learning Research},
  volume={16},
  number={1},
  pages={2147--2205},
  year={2015},
  publisher={JMLR. org}
}

@article{mooij2020joint,
  title={Joint causal inference from multiple contexts},
  author={Mooij, Joris M and Magliacane, Sara and Claassen, Tom},
  journal={The Journal of Machine Learning Research},
  volume={21},
  number={1},
  pages={3919--4026},
  year={2020},
  publisher={JMLRORG}
}

@article{shimizu2012joint,
	title={Joint estimation of linear non-Gaussian acyclic models},
	author={Shimizu, Shohei},
	journal={Neurocomputing},
	volume={81},
	pages={104--107},
	year={2012},
	publisher={Elsevier}
}

@article{Wang2020joint,
	title = {High-dimensional joint estimation of multiple directed Gaussian graphical models},
	author = {Yuhao Wang and Santiago Segarra and Caroline Uhler},
	volume = {14},
	journal = {Electronic Journal of Statistics},
	number = {1},
	pages = {2439 -- 2483},
	year = {2020}
}

@article{chen2021multi,
	title={Multi-task learning of order-consistent causal graphs},
	author={Chen, Xinshi and Sun, Haoran and Ellington, Caleb and Xing, Eric and Song, Le},
	journal={Advances in Neural Information Processing Systems},
	volume={34},
	pages={11083--11095},
	year={2021}
}

@article{zhuang2020comprehensive,
	title={A comprehensive survey on transfer learning},
	author={Zhuang, Fuzhen and Qi, Zhiyuan and Duan, Keyu and Xi, Dongbo and Zhu, Yongchun and Zhu, Hengshu and Xiong, Hui and He, Qing},
	journal={Proceedings of the IEEE},
	volume={109},
	number={1},
	pages={43--76},
	year={2020},
	publisher={IEEE}
}

@article{cai2021transfer,
	title={Transfer learning for nonparametric classification: Minimax rate and adaptive classifier},
	author={Cai, T Tony and Wei, Hongji},
	journal={The Annals of Statistics},
	volume={49},
	number={1},
	pages={100--128},
	year={2021},
	publisher={Institute of Mathematical Statistics}
}

@article{li2022transfera,
	title={Transfer learning in large-scale Gaussian graphical models with false discovery rate control},
	author={Li, Sai and Cai, T Tony and Li, Hongzhe},
	journal={Journal of the American Statistical Association},
	volume={118},
	number={543},
	pages={2171-2183},
	year={2023},
	publisher={Taylor \& Francis}
}

@article{li2022transferb,
	title={Transfer learning for high-dimensional linear regression: Prediction, estimation, and minimax optimality},
	author={Li, Sai and Cai, T Tony and Li, Hongzhe},
	journal={Journal of the Royal Statistical Society: Series B (Statistical Methodology)},
	volume={84},
	number={1},
	pages={149–173},
	year={2022},
	publisher={Wiley Online Library}
}

@article{tian2022transfer,
	title={Transfer learning under high-dimensional generalized linear models},
	author={Tian, Ye and Feng, Yang},
	journal={Journal of the American Statistical Association},
	volume={118},
	number={544},
	pages={2684-2697},
	year={2023},
	publisher={Taylor \& Francis}
}

@article{li2023estimation,
	title={Estimation and inference for high-dimensional generalized linear models with knowledge transfer},
	author={Li, Sai and Zhang, Linjun and Cai, T Tony and Li, Hongzhe},
	journal={Journal of the American Statistical Association},
	volume={119},
	number={546},
	pages={1274-1285},
	year={2024},
	publisher={Taylor \& Francis}
}

@article{li2021targeting,
	title={Targeting underrepresented populations in precision medicine: A federated transfer learning approach},
	author={Li, Sai and Cai, Tianxi and Duan, Rui},
	journal={The Annals of Applied Statistics},
	volume={17},
	number={4},
	pages={2970-2992},
	year={2024}
}

@article{zhao2022learning,
	title={Learning linear non-Gaussian directed acyclic graph with diverging number of nodes},
	author={Zhao, Ruixuan and He, Xin and Wang, Junhui},
	journal={Journal of Machine Learning Research},
	volume={23},
	number={269},
	pages={1--34},
	year={2022}
}

@inproceedings{ghoshal2018learning,
	title={Learning linear structural equation models in polynomial time and sample complexity},
	author={Ghoshal, Asish and Honorio, Jean},
	booktitle={International Conference on Artificial Intelligence and Statistics},
	pages={1466--1475},
	year={2018},
	organization={PMLR}
}

@article{liu2015fast,
	title={Fast and adaptive sparse precision matrix estimation in high dimensions},
	author={Liu, Weidong and Luo, Xi},
	journal={Journal of Multivariate Analysis},
	volume={135},
	pages={153--162},
	year={2015},
	publisher={Elsevier}
}

@article{szekely2007measuring,
	title={Measuring and testing dependence by correlation of distances},
	author={Sz{\'e}kely, G{\'a}bor J and Rizzo, Maria L and Bakirov, Nail K},
	journal={The Annals of Statistics},
	volume={35},
	number={6},
	pages={2769--2794},
	year={2007}
}

@article{zhang2014sparse,
	title={Sparse precision matrix estimation via lasso penalized D-trace loss},
	author={Zhang, Teng and Zou, Hui},
	journal={Biometrika},
	volume={101},
	number={1},
	pages={103--120},
	year={2014},
	publisher={Oxford University Press}
}

@article{lam2009sparsistency,
	title={Sparsistency and rates of convergence in large covariance matrix estimation},
	author={Lam, Clifford and Fan, Jianqing},
	journal={The Annals of Statistics},
	volume={37},
	number={6B},
	pages={4254--4278},
	year={2009},
	publisher={Institute of Mathematical Statistics}
}

@article{ravikumar2011high,
	title={High-dimensional covariance estimation by minimizing $\ell_1$-penalized log-determinant divergence},
	author={Ravikumar, Pradeep and Wainwright, Martin J and Raskutti, Garvesh and Yu, Bin},
	journal={Electronic Journal of Statistics},
	volume={5},
	pages={935--980},
	year={2011},
	publisher={Institute of Mathematical Statistics and Bernoulli Society}
}

@article{kalisch2007estimating,
	title={Estimating high-dimensional directed acyclic graphs with the PC-algorithm.},
	author={Kalisch, Markus and B{\"u}hlman, Peter},
	journal={Journal of Machine Learning Research},
	volume={8},
	number={3},
	year={2007}
}

@article{hyvarinen2013pairwise,
	title={Pairwise likelihood ratios for estimation of non-Gaussian structural equation models},
	author={Hyv{\"a}rinen, Aapo and Smith, Stephen M},
	journal={The Journal of Machine Learning Research},
	volume={14},
	number={1},
	pages={111--152},
	year={2013},
	publisher={JMLR.org}
}

@article{tsamardinos2006max,
	title={The max-min hill-climbing Bayesian network structure learning algorithm},
	author={Tsamardinos, Ioannis and Brown, Laura E and Aliferis, Constantin F},
	journal={Machine learning},
	volume={65},
	pages={31--78},
	year={2006},
	publisher={Springer}
}

@article{guo2018global,
	title={Global characterization of T cells in non-small-cell lung cancer by single-cell sequencing},
	author={Guo, Xinyi and Zhang, Yuanyuan and Zheng, Liangtao and Zheng, Chunhong and Song, Jintao and Zhang, Qiming and Kang, Boxi and Liu, Zhouzerui and Jin, Liang and Xing, Rui and others},
	journal={Nature medicine},
	volume={24},
	number={7},
	pages={978--985},
	year={2018},
	publisher={Nature Publishing Group US New York}
}

@article{sun2016long,
	title={Long non-coding RNA NEAT1 promotes non-small cell lung cancer progression through regulation of miR-377-3p-E2F3 pathway},
	author={Sun, Chengcao and Li, Shujun and Zhang, Feng and Xi, Yongyong and Wang, Liang and Bi, Yongyi and Li, Dejia},
	journal={Oncotarget},
	volume={7},
	number={32},
	pages={51784},
	year={2016}
}

@article{lee2011roles,
	title={Roles of AKT1 and AKT2 in non-small cell lung cancer cell survival, growth, and migration},
	author={Lee, Myoung W and Kim, Dae S and Lee, Joo H and Lee, Bum S and Lee, Soo H and Jung, Hye L and Sung, Ki W and Kim, Heung T and Yoo, Keon H and Koo, Hong H},
	journal={Cancer science},
	volume={102},
	number={10},
	pages={1822--1828},
	year={2011},
	publisher={Wiley Online Library}
}

@article{yang2023plcg2,
	title={PLCG2 can exist in eccDNA and contribute to the metastasis of non-small cell lung cancer by regulating mitochondrial respiration},
	author={Yang, Yongfeng and Yang, Ying and Huang, Hong and Song, Tingting and Mao, Shengqiang and Liu, Dan and Zhang, Li and Li, Weimin},
	journal={Cell death \& disease},
	volume={14},
	number={4},
	pages={257},
	year={2023},
	publisher={Nature Publishing Group UK London}
}

@article{zhang2018identification,
	title={Identification of the tumor-suppressive function of circular RNA FOXO3 in non-small cell lung cancer through sponging miR-155},
	author={Zhang, Yanni and Zhao, Hui and Zhang, Lichuan},
	journal={Molecular Medicine Reports},
	volume={17},
	number={6},
	pages={7692--7700},
	year={2018},
	publisher={Spandidos Publications}
}

@article{shao2014association,
	title={Association of POLK polymorphisms with platinum-based chemotherapy response and severe toxicity in non-small cell lung cancer patients},
	author={Shao, Minhua and Jin, Bo and Niu, Yanjie and Ye, Junyi and Lu, Daru and Han, Baohui},
	journal={Cell biochemistry and biophysics},
	volume={70},
	number={2},
	pages={1227--1237},
	year={2014},
	publisher={Springer}
}

@inproceedings{zhao2018non,
	title={Non-Small-Cell Lung Cancer Gene Expression Network Inference Using Gaussian Graphic Models},
	author={Zhao, Haitao and Duan, Zhong-Hui},
	booktitle={Proceedings of the International Conference on Bioinformatics \& Computational Biology (BIOCOMP)},
	pages={31--37},
	year={2018}
}

@article{chin2014targeting,
	title={Targeting Akt3 signaling in triple-negative breast cancer},
	author={Chin, Y Rebecca and Yoshida, Taku and Marusyk, Andriy and Beck, Andrew H and Polyak, Kornelia and Toker, Alex},
	journal={Cancer research},
	volume={74},
	number={3},
	pages={964--973},
	year={2014},
	publisher={American Association for Cancer Research}
}

@article{zhang2024upregulation,
	title={Upregulation of FGF9 and NOVA1 in cancer-associated fibroblasts promotes cell proliferation, invasion and migration of triple negative breast cancer},
	author={Zhang, Bo and Liu, Yang and Yu, Jinsong and Lin, Xi},
	journal={Drug Development Research},
	volume={85},
	number={3},
	pages={e22185},
	year={2024},
	publisher={Wiley Online Library}
}

@article{li2021hes1,
	title={HES1 promotes breast cancer stem cells by elevating Slug in triple-negative breast cancer},
	author={Li, Xiaoying and Li, Yang and Du, Xianqiang and Wang, Xu and Guan, Shu and Cao, Yu and Jin, Feng and Li, Feng},
	journal={International journal of biological sciences},
	volume={17},
	number={1},
	pages={247},
	year={2021}
}

@article{kim2016association,
	title={Association between mutation and expression of TP53 as a potential prognostic marker of triple-negative breast cancer},
	author={Kim, Ji-Yeon and Park, Kyunghee and Jung, Hae Hyun and Lee, Eunjin and Cho, Eun Yoon and Lee, Kwang Hee and Bae, Soo Youn and Lee, Se Kyung and Kim, Seok Won and Lee, Jeong Eon and others},
	journal={Cancer research and treatment: official journal of Korean Cancer Association},
	volume={48},
	number={4},
	pages={1338--1350},
	year={2016},
	publisher={Korean Cancer Association}
}

@article{pohl2017wnt,
	title={Wnt signaling in triple-negative breast cancer},
	author={Pohl, S{\"O}-G and Brook, N and Agostino, Mark and Arfuso, Frank and Kumar, Alan Prem and Dharmarajan, Arunasalam},
	journal={Oncogenesis},
	volume={6},
	number={4},
	pages={e310--e310},
	year={2017},
	publisher={Nature Publishing Group}
}

\end{document}